\documentclass{article}

\usepackage{arxiv}

\usepackage{caption}
\usepackage[
backend=bibtex,
style=numeric, 
citestyle=ieee,
sorting=none,
]{biblatex}
\usepackage{amsmath, amssymb, latexsym}
\usepackage{tikz}
\usetikzlibrary{positioning, calc, decorations.pathreplacing}
\usetikzlibrary{decorations.pathreplacing}
\usetikzlibrary{fadings}
\usepackage{algorithm}
\usepackage{algcompatible}

\newcommand{\supp}{\mathrm{supp}}

\algnewcommand\algorithmicto{\textbf{to}}
\algnewcommand\RETURN{\State \textbf{return} }
\algnewcommand\And{\textbf{and} }
\algnewcommand{\AND}{\algorithmicand}

\usepackage{bm}

\DeclareMathOperator*{\argmax}{arg\,max}

\usepackage{amsmath,amsfonts,amsthm} 
\usepackage{amssymb}
\usepackage{blkarray}
\usepackage{bm}
\usepackage{mathtools}
\usepackage{empheq} 

\DeclareMathOperator{\E}{\mathbb{E}}

\newtheorem{definition}{Definition}[section]

\usepackage{mwe}
\usepackage{subfig}

\usepackage[utf8]{inputenc} 
\usepackage[T1]{fontenc}    
\usepackage{hyperref}       
\usepackage{url}            
\usepackage{booktabs}       
\usepackage{amsfonts}       
\usepackage{nicefrac}       
\usepackage{microtype}      
\usepackage{cleveref}       
\usepackage{lipsum}         
\usepackage{graphicx}
\usepackage{doi}

\def\keywordname{{\bfseries \emph Keywords}}%
\def\keywords#1{\par\addvspace\medskipamount{\rightskip=0pt plus1cm
		\def\and{\ifhmode\unskip\nobreak\fi\ $\cdot$
		}\noindent\keywordname\enspace\ignorespaces#1\par}}

\usepackage{tikz}
\usetikzlibrary{decorations.pathreplacing}
\usetikzlibrary{fadings}
\usetikzlibrary{shapes,arrows,positioning}
\usetikzlibrary{chains}
\usetikzlibrary{arrows.meta,shadows}
\usepackage{tabularx}
\usepackage{longtable}

\tikzset{
	frame/.style={	
		rectangle, draw,
		text width=6em, text centered,
		minimum height=4em,drop shadow,fill=white,
		rounded corners,
	},
	line/.style={
		draw, -{Latex},rounded corners=3mm,
	}	
}

\usepackage{algorithm}
\usepackage{algcompatible}
\usepackage{algpseudocode}

\algnewcommand\algorithmicto{\textit{to}}
\algnewcommand\RETURN{\State \textit{return} }
\algnewcommand\And{\textit{and} }
\algnewcommand{\AND}{\algorithmicand}

\usepackage{listings}

\usepackage{pythonhighlight}
\title{Advancing Investment Frontiers: Industry-grade Deep Reinforcement Learning for Portfolio Optimization}

\newif\ifuniqueAffiliation
\uniqueAffiliationtrue
\date{February 2024}

\makeatletter
\def\@fnsymbol#1{}
\makeatother

\ifuniqueAffiliation 
\author{Philip Ndikum \thanks{Disclaimer: This research paper, authored by Philip Ndikum \& Serge Ndikum, is for informational and academic purposes only. The authors disclaim any representation or warranty for its accuracy or completeness. It is not intended as investment advice or an endorsement of any specific investment or strategy. Copyright \copyright \; 2024, Philip Ndikum \& Serge Ndikum. All rights reserved.
	} \\
	\And
	Serge Ndikum \\
}
\else
\usepackage{authblk}

\newbox{\orcid}\sbox{\orcid}{\includegraphics[scale=0.06]{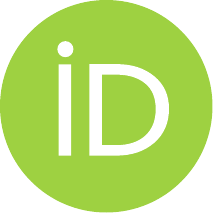}} 
\author[1]{%
	Philip N\thanks{\href{mailto:EMAIL@fas.harvard.edu}{\texttt{EMAIL@fas.harvard.edu}}. This research paper is intended for informational and academic purposes only, and should not be construed as investment, legal, or financial advice. Copyright \copyright \; 2024 Philip Ndikum and Serge Ndikum. All rights reserved.}%
}
\author[2]{%
	Serge N%
}
\affil[1]{Department of [Your Department], Harvard University, Cambridge, MA 02138}
\affil[2]{Department of [Your Department], Northwestern University, Evanston, IL 60208}
\fi

\renewcommand{\headeright}{}
\renewcommand{\undertitle}{}
\renewcommand{\shorttitle}{Copyright \copyright \; 2024 Philip Ndikum \& Serge Ndikum. All rights reserved.}

\hypersetup{
pdftitle={Advancing Investment Frontiers: Industry-grade Reinforcement Learning for Portfolio Optimization},
pdfsubject={q-bio.NC, q-bio.QM},
pdfauthor={Philip Ndikum, Serge Ndikum},
pdfkeywords={First keyword, Second keyword, More},
}

\begin{document}
\maketitle

\begin{abstract}
This research paper delves into the application of Deep Reinforcement Learning (DRL) in asset-class agnostic portfolio optimization, integrating industry-grade methodologies with quantitative finance. At the heart of this integration is our robust framework that not only merges advanced DRL algorithms with modern computational techniques but also emphasizes stringent statistical analysis, software engineering and regulatory compliance. To the best of our knowledge, this is the first study integrating financial Reinforcement Learning with sim-to-real methodologies from robotics and mathematical physics, thus enriching our frameworks and arguments with this unique perspective. Our research culminates with the introduction of AlphaOptimizerNet, a proprietary Reinforcement Learning agent (and corresponding library). Developed from a synthesis of state-of-the-art (SOTA) literature and our unique interdisciplinary methodology, AlphaOptimizerNet demonstrates encouraging risk-return optimization across various asset classes with realistic constraints. These preliminary results underscore the practical efficacy of our frameworks. As the finance sector increasingly gravitates towards advanced algorithmic solutions, our study bridges theoretical advancements with real-world applicability, offering a template for ensuring safety and robust standards in this technologically driven future.
\end{abstract}
\keywords{Portfolio Optimization, Artificial Intelligence, Mathematical Finance, Deep Reinforcement Learning, Robotics, Sim-to-Real Transfer, Risk Management, Regulation, Software Engineering System Design.}

\section{Introduction and Broader Impact}

The integration of Reinforcement Learning (RL) in asset-class agnostic portfolio optimization marks the beginning of a transformative era in quantitative finance, demanding interdisciplinary expertise and novel approaches. This study exemplifies this transformative shift, introducing an industry-grade RL framework integrated with state-of-the-art (SOTA) computational techniques, all tailored for real-world financial applications. To our knowledge, this is the first research paper to adapt simulation-to-real (sim-to-real) transfer perspectives from robotics and mathematical physics to finance. Our frameworks are designed to navigate high-stakes, non-stationary market dynamics within stringent regulatory frameworks. Arguably, this multidisciplinary approach enhances the practicality and robustness of financial models, paving the way for future research that rigorously accounts for historical risks \cite{ndikum2020machine, wilmott2009quants,bailey2021finance, tung2011financialrisk, guhr2015nonstationarity, al2021optimization, de201810, , millea2021deep, andersen2014regulation, makridis2018rise}. The domain of Deep Reinforcement Learning (DRL) has seen significant breakthroughs across various complex fields, underlining its potential to tackle previously insurmountable challenges. Among these, the success of DeepMind's AlphaGo Zero in mastering the ancient game of Go stands as a testament to RL's capability in navigating intricate strategic landscapes. Reinforcement Learning's effectiveness in complex games, particularly in non-stationary environments, with partial information and multi-agent dynamics, closely mirrors the intricate challenges faced in financial markets. Recent advancements in RL, especially in diplomatic simulation games, underscore its potential in navigating stochastic decision-making processes under uncertainty and incomplete information. Such developments are emblematic of RL's adaptability and skill, mirroring the complexities of financial portfolio management and signifying the technology's applicability in this sophisticated domain \cite{holcomb2018overview, heinrich2016deep, paquette2019no, brown2020combining, yao2020solving, bakhtin2021no}. In robotics, leading companies like Boston Dynamics, DiDi, Tesla and others are leveraging Reinforcement Learning to develop advanced autonomous robots and vehicles. This highlights RL's versatility and effectiveness in various high-stakes scenarios \cite{kiran2021deep,jagannath2021deep, chahine2023robust, qin2020ride}. In a parallel development, Large Language Models (LLMs) have undergone significant evolution through Reinforcement Learning with Human Feedback (RLHF). These models have not only achieved progress in natural language processing but also show promise in specialized domains like finance. Preliminary research suggests that financial LLMs may eventually outperform traditional human analysts in analyzing and interpreting complex financial data, indicating a potential shift in how financial information is processed and utilized \cite{sejnowski2023large, uchendu2021turingbench, peng2023instruction, wu2023bloomberggpt}.

 Our research endeavors to bridge the gap between theoretical concepts in academia and the dynamic, often unpredictable realities of real-world financial markets. Despite commendable and ambitious goals, many financial RL papers and open-source software artifacts, such as FinRL and its derivatives, often do not fully meet the stringent requirements of robust software engineering, regulatory compliance, and practical market constraints \cite{liu2020finrl, liu2022finrl, li2021finrl, song2022safe}. We hope that this paper offers insights to enhance industry-grade applications of Reinforcement Learning in portfolio optimization and Financial RL more broadly. Our collaboration was born from our observation that many available open-source tools lack consistent requirements, robust documentation, and often exhibit coding errors stemming from inadequate finance domain knowledge. It is worth noting, that the arguments presented also align with industrial trends in the Artificial Intelligence (AI) industry, with many practitioners advocating for improved engineering  and software design practices \cite{stable-baselines3, huang2022cleanrl, towers_gymnasium_2023, weng2022envpool, nguyen2023towards, zaharia2024compoundsystems}. In economics and AI research, replication challenges and incomplete findings, have been addressed in many excellent papers by industry practitioners but few have provided frameworks to concretely address these challenges in modern finance \cite{jensen2023there, jensen2021there, hou2020replicating, echtler2018open, hutson2018artificial, gibney2022ai, gueheneuc2019empirical, hullman2022worst, tong2019statistical}. Given the high stakes of financial management, as evidenced by past financial crises and the fluctuating performance of investors, it is crucial that RL algorithms deployed in finance undergo rigorous stress-testing, statistical experiment design and conform to high engineering standards \cite{mead2012statistical, parkinson2021we, schiefer2021statistical, kang2020design, munger2021statistical}. The proprietary library and RL agent, \textit{AlphaOptimizerNet}, were meticulously crafted to ensure robustness in experiment design and performance analysis\footnote{It is important to note that many developments in the fields of finance and investment are proprietary and thus not publicly disclosed. As a result, this paper solely references publicly available academic literature and does not speculate nor comment on the advancements or strategies employed within private institutions. The development of our RL agent, \textit{AlphaOptimizerNet}, and the accompanying proprietary library, is an example of such non-public, specialized work in the field.}. Embracing an iterative and system-driven design philosophy, the development acknowledges the nuanced complexities inherent in contemporary AI system design. This research underscores the potential significance of sim-to-real considerations, a central theme throughout this paper. Such an approach may be crucial for organizations operating in an ecosystem increasingly shaped by advanced algorithms. Our objective is to argue for the adoption of elevated standards in both academic and industrial realms, contributing to the future landscape of financial management. The insights presented in this research might be particularly relevant to sovereign wealth funds, family offices, pension funds and public policy makers. In today's financial environment, these institutions play a pivotal role as guardians of capital for the benefit of families, citizens, and nations. We aim to enhance individual well-being and positively influence the broader trajectory towards global prosperity.

\section{Transcending Traditional Portfolio Optimization: A Reinforcement Learning Approach}

\subsection{Modern Portfolio Theory: Limitations and RL's Role in Future Directions}

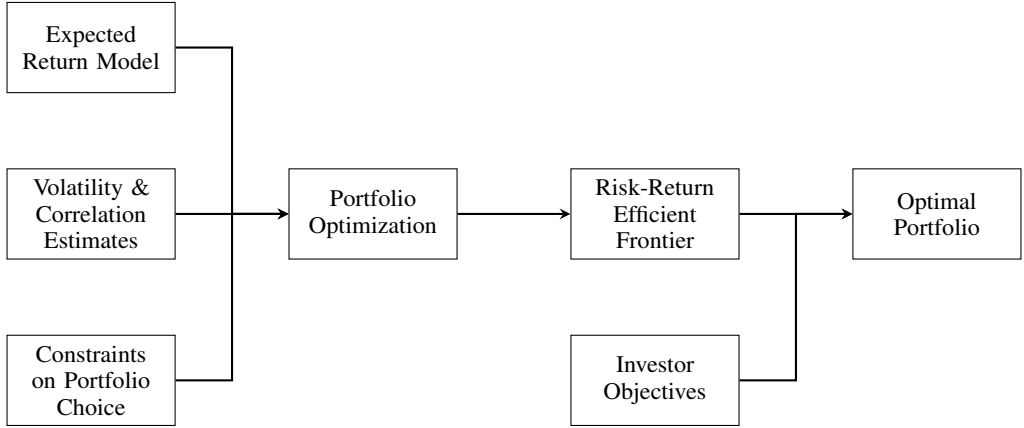
\begin{figure}[htp!]
	\centering
	\begin{tikzpicture}[
		block/.style={rectangle, draw, text width=2cm, text centered, minimum height=1.2cm, font=\footnotesize},
		arrow/.style={thick,->,>=stealth}
		]
		
		\node[block] (expected_return) {Expected Return Model};
		\node[block, below=of expected_return] (volatility_correlation) {Volatility \& Correlation Estimates};
		\node[block, below=of volatility_correlation] (constraints) {Constraints on Portfolio Choice};
		
		\node[block, right=1.5cm of volatility_correlation] (portfolio_optimization) {Portfolio Optimization};
		
		\draw[arrow] (expected_return.east) -- ++(0.75, 0) |- (portfolio_optimization.west);
		\draw[arrow] (volatility_correlation.east) -- (portfolio_optimization.west);
		\draw[arrow] (constraints.east) -- ++(0.75, 0) |- (portfolio_optimization.west);
		
		\node[block, right=1.5cm of portfolio_optimization] (efficient_frontier) {Risk-Return Efficient Frontier};
		\node[block, below=of efficient_frontier] (investor_objectives) {Investor Objectives};
		
		\draw[arrow] (portfolio_optimization) -- (efficient_frontier);
		
		\node[block, right=1.5cm of efficient_frontier] (optimal_portfolio) {Optimal Portfolio};
		
		\draw[arrow] (efficient_frontier.east) -- (optimal_portfolio.west);
		\draw[arrow] (investor_objectives.east) -- ++(0.75, 0) |- (optimal_portfolio.west);
		
	\end{tikzpicture}
	\caption[Quantitative Investing Process in Modern Portfolio Theory]{
		This diagram, adapted from Fabozzi et al. (2002) \cite{fabozzi2002legacy}, succinctly outlines the Modern Portfolio Theory (MPT) investing process. It shows the progression from expected return and volatility \& correlation modeling, through the inclusion of portfolio constraints, to optimization, culminating in the establishment of the risk-return efficient frontier and selection of the optimal portfolio.}
	\label{fig:quant_investing_process}
\end{figure}

Portfolio optimization is a fundamental component of quantitative investing, tasked with the strategic allocation of assets to maximize returns while minimizing risk. It encompasses a broad spectrum of asset classes and investment timelines, reflecting the multifaceted nature of financial markets. In our research, we consider portfolio optimization as a prime example to demonstrate the application of our Deep Reinforcement Learning (RL) methodologies. Building on the work of Ndikum (2020) \cite{ndikum2020machine}, we highlight the critical role of financial, econometric and regulatory  domain knowledge for the successful deployment of advanced AI algorithms in finance. As we delve into Deep RL techniques for portfolio optimization, we consciously choose to employ simplified mathematical formulations. This approach, aimed at making our work accessible to a diverse audience from various fields, reflects our commitment to balance technical detail with general comprehensibility. Through this strategy, we aim not only to engage a broader readership but also to foster meaningful discourse across different disciplines, maintaining the essence and integrity of our arguments throughout \footnote{For a comprehensive textbook on Reinforcement Learning, we  recommend \textit{Reinforcement Learning and Stochastic Optimization: A unified framework for sequential decisions} by Professor Warren Powell, of Princeton University, 2022. \cite{powell2022reinforcement}.}.   Our objectives in this section are twofold: Firstly, we aim to elucidate the fundamental concept of diversification and the mechanics underlying Modern Portfolio Theory (MPT). Secondly, we delve into the inherent limitations of MPT, particularly in the context of today's dynamic financial markets. A thorough grasp of these foundational concepts is indispensable, as it lays the groundwork for a deeper appreciation of the sim-to-real framework presented later in our paper. In the realm of portfolio management, diversification is not merely a strategy for risk mitigation; instead it fundamentally forms the philosophical underpinning of Modern Portfolio Theory (MPT). At the time of its inception, MPT represented a paradigm shift in financial portfolio management, systematically harnessing the principles of diversification to mathematically balance risk and return. This innovative approach revolutionized the way portfolios were constructed, moving beyond intuitive strategies to a quantifiable and objective models for optimization.  Diversification, a central tenet of Modern Portfolio Theory (MPT), can be quantitatively expressed by considering the expected risk, represented as the variance of returns \(\sigma_p^2\), in a diversified portfolio of \(n\) distinct assets. This approach to risk quantification considers the expected fluctuations in portfolio returns. We follow a mathematical formulation similar to Bodie et al. (2020) \cite{bodie2020investments}:
\begin{equation}
	\sigma_p^2 = \sum_{i=1}^{n} \sum_{j=1}^{n} w_i w_j \text{Cov}(r_i, r_j).
	\label{eqn:portfolio_variance}
\end{equation}
Equation \ref{eqn:portfolio_variance} calculates the expected variance of portfolio returns, \(\sigma_p^2\), illustrating how diversification impacts this risk\footnote{In this context, the notation \( \sigma_p^2 \) represents the variance of the portfolio's returns, and \( \rho_{ij} \) (Greek Rho) refers to the correlation coefficient between the returns of assets \( i \) and \( j \).}. Here, \(w_i\) and \(w_j\) represent the portfolio weights allocated to assets \(i\) and \(j\) respectively, and \(\text{Cov}(r_i, r_j)\) is the covariance between their returns, measuring how their returns move in relation to each other. The essence of diversification in risk management lies in combining assets with varying degrees of correlation, thereby reducing the portfolio's overall expected risk. The influence of this diversification becomes more significant as the number of assets increases and their correlations diversify. This dynamic is captured in the following equation:
\begin{equation}
	\lim_{{n \to \infty, \atop \rho_{ij} \to 0}} \sigma_p^2 = \lim_{{n \to \infty, \atop \rho_{ij} \to 0}} \sum_{i=1}^{n} \sum_{j=1}^{n} w_i w_j \text{Cov}(r_i, r_j) = C = \inf_{i \in \mathcal{N}} \sigma_i^2,
	\label{eqn:diversification_impact}
\end{equation}
Equation \ref{eqn:diversification_impact} captures the theoretical impact of diversification on the expected risk of a portfolio, measured using the portfolio variance \(\sigma_p^2\). As the number of assets \(n\) in the portfolio grows indefinitely and their pairwise correlations \(\rho_{ij}\) converge to zero, the portfolio's expected risk trends towards a theoretical minimum limit, denoted as \(C\). This limit, \(C\), is specifically defined as the infimum (\(\inf\)) of the variances (\(\sigma_i^2\)) of the individual assets within the portfolio set \(\mathcal{N}\). In essence, \(C\) represents the lowest level of risk (variance) that is theoretically achievable under the ideal scenario of infinite diversification with completely uncorrelated assets.  This concept is central to portfolio theory, emphasizing the advantage of having a diverse mix of uncorrelated or negatively correlated assets in a portfolio. Such diversification is a key strategy in risk management, aiming to reduce the impact of market volatility and improve the overall stability of an investment portfolio. In practical terms, it means spreading investments across a variety of asset classes to buffer against unexpected market movements, thereby enhancing the resilience of one's financial portfolio. The use of infimum notation in this equation highlights the inherently positive nature of this theoretical minimum risk. It underscores that even with ideal diversification, portfolio risk doesn't reduce to zero but converges towards the lowest variance found among the individual assets.

This concept of minimizing expected risk through diversification leads us to the core principles of Modern Portfolio Theory (MPT), as formulated by Harry Markowitz. In recognition of their pivotal contribution to the field, Markowitz, along with Merton Miller and William Sharpe, were collectively awarded the Nobel Prize in Economics in 1990 \cite{markowitz1967portfolio, modigliani1958cost, sharpe1964capital, markowitz1991foundations}. MPT pragmatically applies the principle of diversification in the context of portfolio optimization. MPT algorithms are often categorized under Quadratic Programming (QP) or Mixed Integer Programming (MIP) problems within the fields of numerical analysis and applied mathematics. Definition \ref{def:mpt_unconstrained} provides a mathematical formulation of a single-period MPT, drawing upon the formulations by Chang et al. (2002) and Cesarone et al. (2013) \cite{chang2000heuristics, cesarone2013new, kouzoupis2018recent, hirschberger2010large, boyd2017multi}. In this definition, \( n \) represents the total number of financial assets, \( w_i \) represents the portfolio allocation to asset \( i \) in the range \( [0, 1] \). The expected return and covariance for each asset are denoted as \( \mu_i \) and \( \sigma_{ij} \), respectively. In simple terms, we aim to optimize portfolio weights to minimize risk while targeting a specific expected return \( \E[R] = R \). This unconstrained formulation can be solved efficiently with modern computational tools, allowing for the calculation of optimal portfolio weights, designated as \(\phi(\E[R])\). The resultant \textit{efficient frontier} identifies portfolios that optimize expected return for various levels of risk within the interval \([R_{\text{min}}, R_{\text{max}}]\). In real-world investing scenarios, this deceptively simple formulation of Modern Portfolio Theory (MPT) is significantly complicated by the introduction of various real-world constraints.
\begin{definition}[Markowitz Single Period Unconstrained Portfolio Optimization\cite{chang2000heuristics, cesarone2013new}]
	\begin{equation}
		\begin{aligned}
			\min \quad & \sum_{i=1}^{n} \sum_{j=1}^{n}\sigma_{ij} w_i w_j  \\
			\textrm{s.t.} \quad & \sum_{i=1}^{n} w_i \mu_i = \E[R] \\
			& \sum_{i=1}^{n} w_i = 1 \\
			& w_i \in [0,1], \text{ for } i \in \{1, \dots, n\}.
		\end{aligned}
		\label{eqn:mpt_unconstrained}
	\end{equation}
	\label{def:mpt_unconstrained}
\end{definition}
The Financial literature published in the last few decades has thoroughly explored and addressed these constraints, adding complexity to MPT models. The intricate and high-stakes nature of finance necessitates a deep dive into this rich history. While contemporary and future technological advancements in Reinforcement Learning, present innovative possibilities, their effective application demands a thorough grounding in the enduring realities of financial markets and regulatory environments. Neglecting the depth of historical financial knowledge can oversimplify intricate market dynamics, potentially steering strategies towards catastrophic outcomes for financial institutions and investors in both the private and public sectors. Below, we enumerate some constraints and place them in context with real-world problems:
\begin{enumerate}
	\item \textbf{Cardinality Constraint:}
	This constraint limits the number of assets in a portfolio to a maximum of \(n\):
	\begin{equation}
		|\supp(w)|_0 \leq n, \quad \text{where } \supp(w) := \{i: w_i > 0\}.
	\end{equation}
	Here, \(\supp(w)\) denotes the set of indices \(i\) for which the investment in asset \(i\) (\(w_i\)) is greater than zero, implying active inclusion of the asset in the portfolio. The expression \(|\supp(w)|_0\) counts the total number of distinct assets with non-zero investments in the portfolio. The constraint \(|\supp(w)|_0 \leq n\) ensures that this number does not exceed the predetermined maximum \(n\), thereby limiting the total number of different assets included in the portfolio. In asset management, this constraint is relevant for ensuring portfolio transparency and manageability.

	\item \textbf{Volume Constraint:} 
	The volume constraint restricts the allocation to each asset in the portfolio within a specified range, denoted by
	\begin{equation}
		\ell_i \leq w_i \leq u_i \; \text{for each asset } i \in \{1, \dots, n\},
	\end{equation}
	where \(\ell_i\) and \(u_i\) define the lower and upper bounds, respectively, for the investment in asset \(i\). This constraint is critical in preventing over-concentration in a single asset, thereby reducing the risk of significant loss from adverse movements in that asset's value. In high frequency trading (HFT), the volume constraint indirectly influences turnover and liquidity, both of which are critical factors affecting the overall effectiveness of trading strategies.
	
	\item \textbf{Regulatory Constraints and Transaction Costs:} 
	Legal and regulatory constraints may dictate investment limitations in certain sectors or geographies. For example, contemporary investment strategies may be influenced by client or government mandates that encourage investments in certain sectors, such as renewable energy, reflecting broader objectives that may not be captured mathematically by classical models. These external directives can significantly shape portfolio composition. Beyond regulatory considerations, transaction costs, including taxes and brokerage fees, play a crucial role. The incorporation of such constraints can dramatically alter the performance of back-tested simulations in both Modern Portfolio Theory and newer methodologies such as Reinforcement Learning.
\end{enumerate}
As we can observe, the adoption of Modern Portfolio Theory (MPT) in dynamic financial scenarios necessitates confronting a myriad of complexities. Additionally, we note that these classical solutions operate under static market assumptions and are often geared towards single-period frameworks. This inherent limitation impedes their utility in environments characterized by asymmetric risk profiles\footnote{Post-Modern Portfolio Theory (PMPT) seeks to mitigate challenges associated with asymmetric risk. Nevertheless, it predominantly utilizes conventional optimization methods, which are subject to analogous computational challenges \cite{bourachnikova2012investor, racheva2008post}.}, human irrationality, and complex game-theoretic dynamics. It is within this context that the concept of multi-period optimization solutions have gained recent prominence: 

\begin{definition}[Multi-period Optimization]
	\label{def:multi_period_opt}
	Consider a universe of $n$ assets over a time horizon of $K$ periods. We provide a similar formulation to Lezmi et al. (2022) \cite{lezmi2022multi} and transition to a probabilistic framework to model our multi-period optimization as follows:
	\begin{equation}
		W^* = \argmax_{W \in \Omega} \mathbb{E} \left[ U (W) \,|\, \mathcal{F}_t \right]
		\label{eqn:multi_period_optimization_probabilistic}
	\end{equation}
	where 
	\begin{equation*}
		W = \begin{pmatrix} 
			w_{1,t+1} & w_{1,t+2} & \cdots & w_{1,t+K} \\ 
			w_{2,t+1} & w_{2,t+2} & \cdots & w_{2,t+K} \\ 
			\vdots & \vdots & \ddots & \vdots \\ 
			w_{n,t+1} & w_{n,t+2} & \cdots & w_{n,t+K} 
		\end{pmatrix},
	\end{equation*}
	denotes the matrix of asset allocations across $K$ future time periods, with each $w_{i,t+j}$ being the column vector of portfolio weights allocated to asset $i$ at time period $t+j$. Here, $W^*$ represents the optimal series of allocations. The function $U (W)$ represents the inter-temporal utility function, expressing the investor's preference over varied portfolio allocations. The term $\mathcal{F}_t$ is the filtration, representing the cumulative information available up to time $t$, including past market data and events. The optimization is subject to the condition $W \in \Omega$, where $\Omega$ denotes the sample space of all feasible sequences of portfolio allocations, subject to a set of linear and non-linear constraints. This framework allows for the ranking of allocations based on desirability, with superior utility values indicating more favored allocations, factoring in considerations such as returns, risks, and other investor inclinations.
\end{definition}

The multi-period optimization shown in Definition \ref{def:multi_period_opt} aims to forecast and optimize portfolio weights across several forward-looking timeframes. Despite its theoretical robustness, real-world application of multi-period optimization presents substantial challenges adequately addressed and  summarized by Kolm et al. (2014) in their analysis of portfolio optimization's evolution and challenges:
\begin{quote}
	\textit{“In practice, multi-period models are seldom used. There are several practical reasons for that. First, it is often very difficult to accurately estimate return and risk for multiple periods, let alone for a single period. Second, multi-period models are in general computationally intensive, especially if the universe of assets considered is large. Third, the most common existing multi-period models do not handle real-world constraints \dots For these reasons, practitioners typically use single-period models to rebalance the portfolio from one period to another”} - 60 Years of portfolio optimization: Practical challenges and current trends, Kolm et al. (2014) \cite{kolm201460}.
\end{quote}
This complexity is compounded in practical numerical solutions to MPT, such as Mean-Variance Optimization (MVO). MVO applies the principles of MPT in a quantitative framework to identify the optimal asset allocation for a given risk level. A key requirement in MVO is for the covariance matrix, which represents the relationships between the returns of different assets, to be positive semi-definite. This ensures that the calculated variance of the portfolio, a key risk measure in MPT, is always non-negative. Additionally, the inherent variability or volatility in financial markets adds another layer of complexity. The assumption of stable volatility, a cornerstone of MVO, often does not hold true in real-world scenarios where market conditions can rapidly change. In contrast, Reinforcement Learning (RL) offers a promising alternative. RL thrives in environments characterized by uncertainty and the need for sequential decision-making, aligning well with the dynamic nature of financial markets. Unlike MVO, RL does not require the stringent assumptions about volatility and matrix properties. RL's ability to handle complex, dynamic environments, coupled with its flexibility in integrating real-world constraints, makes it a formidable tool in addressing the challenges highlighted by Kolm et al. \cite{bouchaud2008economics, geambasu2013risk, martellini2008toward, kempf2015portfolio}. State-of-the-art RL algorithms offer the potential to navigate the evolving landscape of risks and large datasets, holding the promise of superior and autonomous risk-adjusted returns. This shift in focus towards RL and similar advanced algorithms, however, must not eclipse the importance of grounding these technologies in the contemporary and historical realities of financial markets and regulatory environments \cite{DAlvia2023, akyildirim2014brief}. Given the stakes involved in managing client capital, it seems imperative for the financial industry, similar to other high-risk sectors where RL is currently being used in industry (e.g., robotics), to establish rigorous standards ensuring the responsible deployment of these algorithms. In this paper, we advocate for a paradigm where the effectiveness of techniques such as RL is gauged not merely by their theoretical prowess, but by their alignment with the dynamic - and highly regulated demands of contemporary international finance which we will explore in the subsequent sections.
\subsection{Reinforcement Learning in Portfolio Optimization: Bridging Theory and Practice}
\begin{figure}[htp!]
	\centering
	\begin{tikzpicture}[font=\small, thick, node distance = 4cm, scale=1.1, every node/.style={scale=1.2}]
		\node [frame] (agent) {RL Agent};
		\node [frame, below=1.2cm of agent] (environment) {Financial Environment};
		\draw[line] (agent) -- ++(3.5,0) |- (environment)
		node[right,pos=0.25,align=left] {action $a_t$};
		\coordinate[left=12mm of environment] (P);
		\draw[thin,dashed] (P|-environment.north) -- (P|-environment.south);
		\pgfmathsetmacro{\Ldist}{4mm}
		\draw[line] ([yshift=-\Ldist]environment.west) --
		([yshift=-\Ldist]environment.west -| P) node[midway,above]{$s_{t+1}$};
		\draw[line,thick] ([yshift=\Ldist]environment.west) -- ([yshift=\Ldist]environment.west
		-|P) node[midway,above]{$r_{t+1}$};
		\draw[line] ([yshift=-\Ldist]environment.west -| P) -- ++ (-12mm-\Ldist,0) |-
		([yshift=\Ldist]agent.west) node[left, pos=0.25, align=right] {state $s_t$};
		\draw[line,thick] ([yshift=\Ldist]environment.west -| P) -- ++ (-12mm+\Ldist,0)
		|- ([yshift=-\Ldist]agent.west) node[right,pos=0.25,align=left] {reward $r_t$};
	\end{tikzpicture}
	\caption[A schematic representation of the RL Agent-Financial Environment interaction in portfolio optimization]{A schematic representation of the interaction between the RL Agent and the financial environment in the context of portfolio optimization. The RL Agent represents the algorithm making portfolio decisions, where the state ($s_t$) denotes the current market conditions and portfolio configuration, the action ($a_t$) corresponds to portfolio adjustment decisions, and the reward ($r_t$) reflects the financial outcome, such as risk-adjusted returns, of these decisions. This diagram illustrates how RL adapts and responds to evolving financial scenarios, thereby optimizing portfolio performance.}
	\label{fig:rl_agent_financial_environment_interaction}
\end{figure}
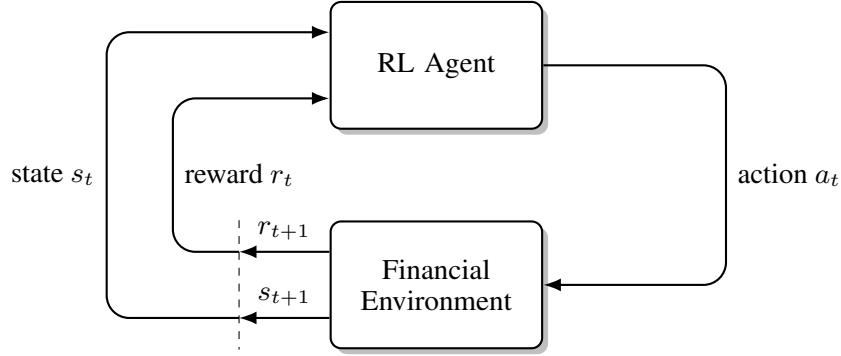
The transition to Reinforcement Learning (RL) in financial portfolio management marks a pivotal shift from conventional strategies to advanced, algorithm-driven approaches. Deep Reinforcement Learning (DRL), an extension of RL, effectively employs deep neural networks to navigate complex, high-dimensional environments. This subject will be further explored in upcoming sections. RL, set apart from the single-period, deterministic models of Modern Portfolio Theory (MPT), adopts a dynamic, data-driven approach. It is adept at managing multiple, even\ \textit{ infinite time horizons} signaling a move towards creating computational equivalents of human traders and investment managers. Reinforcement Learning systems operate as an autonomous agents (advanced algorithm or models), learning and adapting continuously via interactions with a simulated environment that mirrors real-world market complexities. This learning process, entirely driven by data, enables the RL agent to identify patterns and make informed decisions based on extensive market data. This continual, dynamic learning process is essential for developing investment strategies that are responsive to the rapidly changing and unpredictable financial markets. This evolution underscores a departure from the traditional reliance on human emotional resilience in financial decision-making. With its capacity to process large data volumes, RL holds the potential to match or even exceed human capabilities in identifying and capitalizing on market inefficiencies. Free from human psychological biases and fatigue, RL systems introduce a major transformation in the financial sector. They combine human expertise with the precision of computational systems. In this new era, trust in financial decision-making extends beyond human judgment to the dependability and accuracy of these meticulously engineered computational systems. This transition represents a major shift in the landscape of financial portfolio management. To facilitate a deeper understanding and bridge the gap towards our novel sim-to-real frameworks, we employ analogies from robotics throughout our discussion. These analogies not only provide a familiar reference point but also align closely with the principles and challenges inherent in both fields:
\begin{itemize}
\item \textbf{State} (\(s_t\)): In RL, the state at time \(t\), \(s_t\), represents a snapshot of the current situation, analogous to the perception of an environment by an RL agent in robotics. In portfolio optimization, \(s_t\) encapsulates market conditions, including key financial indicators such as asset prices and trading volumes. Understanding the state's formulation requires a nuanced grasp of the asset class and regulatory context, comparable to the detailed environmental awareness necessary for an RL agent in robotics. The strategic design of the state space, similar to sensor range and resolution considerations in robotics, is crucial for capturing detailed market representation while maintaining computational manageability. This process is central to optimizing the RL model's efficiency in interpreting and reacting to market dynamics, paralleling the way an RL agent in robotics processes sensory data to understand its environment.
\item \textbf{Action} (\(a_t\)): The action \(a_t\) at time \(t\) signifies the decision made by the RL agent, comparable to decision-making processes of RL agents in robotics. In portfolio optimization, this action primarily involves determining the asset allocations, represented as a vector. The constraint \(\sum_{i=1}^{n} a_i = 1\) ensures that the total allocation across all assets equals one, reflecting a similar allocation of resources or efforts by RL agents in robotics. This constraint is critical in ensuring that the portfolio weights are proportionally distributed. Actions may also incorporate factors such as maintaining cash reserves or opting for periods of inactivity, dependent on the strategy. The action space design must align with specific investment strategies and comply with regulation and investment mandates, including decisions regarding the use of leverage.
\item \textbf{Reward} (\(r_t\)): The reward \(r_t\) at time \(t\) functions as the guiding incentive for the RL agent, comparable to the objectives pursued in robotics where agents learn to navigate and interact with dynamic and complex physical environments. In portfolio optimization, the agent's reward is typically aligned with the maximization of risk-adjusted returns, tailored to meet investor-specific goals and limitations. This mechanism is crucial in guiding the agent's learning path, enabling it to develop and refine strategies that effectively balance return and risk.
\item \textbf{Environment:} In financial RL, the environment represents the complexities of market dynamics, regulatory frameworks, and economic indicators, similar to the intricate and evolving environments encountered in robotics. The financial environment is inherently non-stationary and only partially observable, posing significant challenges comparable to those faced in robotic navigation and interaction. Designing an RL model for finance requires a deep understanding of market intricacies and nuances. The environment should mirror specific investment timeframes and strategies, with considerations varying greatly between applications such as high-frequency trading and long-term asset management. It's imperative that the environment is realistically modeled to encapsulate market behaviors and constraints, ensuring the RL model's relevance and effectiveness in actual financial scenarios.
\end{itemize}
As we explore Reinforcement Learning (RL) for portfolio optimization, understanding its mathematical foundations is crucial. RL algorithms, though complex, rely on well-established mathematical principles and computational procedures. Creating these systems necessitates expertise in AI, finance, and a deep understanding of regulatory and risk considerations. This skill set is essential for successful deployment of robust Reinforcement Learning systems in finance. Our aim is to present these mathematical concepts clearly and comprehensively, particularly for readers with a finance background. While equations are our focus, the accompanying explanations provide an insight into their underlying principles. Designing RL systems for portfolio optimization requires more than technical proficiency; it entails integrating financial knowledge, AI capabilities, and an understanding of real-world regulations and risks. As we delve into Markov Decision Processes (MDPs), Partially Observable Markov Decision Processes (POMDPs), and key aspects of Deep Learning, our goal is to maintain a balance between technical rigor and accessibility. Subsequent sections will cover mathematical equations while contextualizing them within financial portfolio management.

\subsection{Mathematical Formalism: MDPs, POMDPs, Deep Learning}
\begin{figure}[htp!]
	\centering
	\includegraphics[width=1\textwidth]{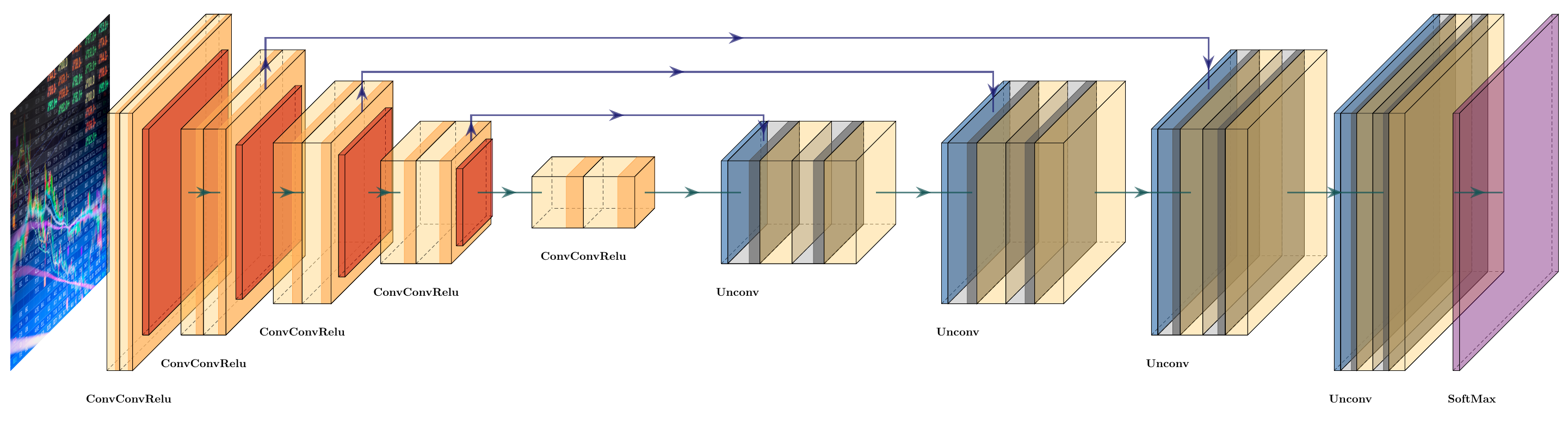}
	\caption[Convolutional Neural Network (U-Net) for Representation Learning]{This figure displays a U-Net, originally a Convolutional Neural Network for biomedical imaging, now adapted to exploit patterns in financial data. Its interoperability underlines the crucial role of interdisciplinary expertise in crafting advanced RL systems for finance.}
	\label{fig:cnn_finance}
\end{figure}
We now delve into the mathematical formalism of Deep Reinforcement Learning (RL) shown in Fig. \ref{fig:rl_agent_financial_environment_interaction}. This section is structured to first introduce intuitive explanations of Markov Decision Processes (MDPs) and Partially Observable Markov Decision Processes (POMDPs), followed by their formal definitions, and then a discussion of Deep Learning and neural networks. A key distinction between broader Reinforcement Learning (RL) approaches and Deep Reinforcement Learning (DRL) is the use of deep neural networks in the latter, enabling them to handle more complex, high-dimensional and noisy financial market data. Although an in-depth technical exploration exceeds the scope of this paper, we introduce neural networks as the key driving force behind our RL algorithms. Analogous to \textit{computational brains}\footnote{It is essential to understand that neural networks, while biologically inspired, do not replicate the full complexity of biological brains. These statistical learning algorithms are abstractions, simplifying key principles from our evolving understanding of neuroscience. Interestingly, there appears to be a correlation between advancements in computational neuroscience and the development of increasingly performant neural network architectures. However, this should not be mistaken for a comprehensive understanding of the brain's intricacies or an implication of neural networks' complete mimicry of biological processes \cite{schmidgall2023brain, sarfraz2023study}.}, these biologically inspired algorithms are indispensable due to their ability as universal function approximators, essential for effective planning, prediction, and decision-making within MDP and POMDP frameworks. In our examination of Markov Decision Processes (MDPs) and Partially Observable Markov Decision Processes (POMDPs), let's offer simplified interpretations before presenting formal definitions. MDPs represent decision-making frameworks with complete market information. Imagine making investment decisions where you have full visibility into asset prices, economic indicators, and market trends. This scenario aligns with an MDP framework. Conversely, POMDPs apply when some market information is hidden or uncertain. Picture a situation where certain variables, such as insider information or future market movements, are unknown. Despite this uncertainty, decisions must be made based on available information. POMDPs provide a framework for decision-making under such conditions, allowing adaptation to market dynamics even with incomplete visibility:

\begin{definition}[Markov Decision Processes in Portfolio Optimization]
	In portfolio optimization, an MDP is characterized by the tuple $(S, A, T, T_0, R, \gamma)$. Here, $S$ represents the array of possible market states, encompassing variables like asset prices and economic indicators. $A$ denotes the range of possible investment actions, reflecting decisions across a portfolio of $n$ assets, thus $A \subseteq [0,1]^n$. The transition function $T : S \times A \rightarrow \Delta(S)$ models the probability of moving from one market state to another, given a specific investment action. $T_0 : S \rightarrow [0, 1]$ defines the initial distribution of states in the market. The reward function $R : S \times A \rightarrow \mathbb{R}$ quantifies the financial impact of each action, considering both returns and associated risks.  Unlike finite-horizon models, the time horizon in this setting is considered infinite, and the discount factor $\gamma \in [0, 1)$ reflects the long-term strategy of the investment approach. The primary objective is to determine an optimal policy $\pi^* : S \rightarrow A$ that maximizes expected returns over this infinite horizon, as formalized in the following equation:
	\begin{equation}
		\label{eqn:mdp_portfolio}
		\pi^* = \arg \max_\pi \mathbb{E}_{s_t, a_t, r_t \sim T, \pi} \left[ \sum_{t=0}^{\infty} \gamma^t r_{t+1} \; \vert \; s_0 \right].
	\end{equation}
	The objective of the RL agent in this framework is to learn the optimal mapping of states to actions (portfolio allocations), thereby maximizing long-term investment returns while adapting to the dynamic nature of financial markets.
\end{definition}

\begin{definition}[Partially Observable Markov Decision Processes in Financial Markets]
	POMDPs extend the MDP framework to scenarios where full knowledge of market states is not directly observable by investors. A POMDP in finance is characterized by the tuple $(S, A, O, T, T_0, R, \gamma)$. Here, $S$ denotes the complete range of market states, $A$ represents the range of investment actions, and $O$ encompasses observable market factors. The transition function $T : S \times A \rightarrow \Delta(S)$ models the dynamics of market state changes following investment actions, and $T_0 : S \rightarrow [0, 1]$ defines the initial state distribution. The observation function $O : S \times A \rightarrow \Delta(O)$ determines the likelihood of observing specific market indicators given the state and action. The reward function $R : S \times A \rightarrow \mathbb{R}$ quantifies the financial impact of actions. The decision-making policy in a POMDP, $\pi : \mathcal{H} \rightarrow A$, where $\mathcal{H}$ is the history of observations, actions, and rewards, depends on both the current and past observations. This history up to time $t$ is represented as $\tau_{0:t} = (o_0, a_0, o_1, r_1, \dots, a_{t-1}, o_t, r_t)$. The objective in a POMDP setting is to identify an optimal policy $\pi^*$ that maximizes expected returns over an infinite horizon, considering both observable and unobservable market factors:
	\begin{equation}
		\label{eqn:pomdp_portfolio}
		\pi^* = \arg \max_\pi \mathbb{E}_{s_t, a_t, o_t, r_t \sim T, O, \pi} \left[ \sum_{t=0}^{\infty} \gamma^t r_{t+1} \; \vert \; o_0 \right].
	\end{equation}
	In financial markets, where complete state information is often unavailable, this model becomes crucial, necessitating decision-making based on partial observations \cite{aastrom1965optimal, littman2009tutorial}.
\end{definition}
In our exploration of decision-making frameworks, such as MDPs and POMDPs within dynamic financial environments, we have encountered foundational concepts that underpin adaptive learning. Now, we pivot our focus to a fundamental concept in Reinforcement Learning (RL) that serves as a linchpin for adaptive decision-making: the Bellman equation. Developed by Richard Bellman in 1957 \cite{bellman1957markovian}, the Bellman equation plays a pivotal role in dissecting decision-making processes into interconnected steps, enabling the evaluation of both immediate rewards and the anticipated utility of future actions. At the heart of RL methodologies lies the concept of the value function, which encapsulates the expected cumulative future rewards an agent can attain from a given state. In the context of portfolio optimization, this value function serves as the cornerstone of the RL agent's learning process, guiding it towards determining the optimal portfolio allocation strategy. By iteratively updating the value function based on observed rewards and transitions between states, the RL agent gradually hones its understanding of the market dynamics and learns to make informed decisions that maximize long-term returns while balancing risk. With this understanding of the value function's significance in RL, let us delve into the formulation of the Bellman equation and its implications for portfolio optimization:
\begin{equation}
	V(s_t) = \max_{a_t} \left[ R(s_t, a_t) + \gamma \sum_{t'=t+1}^{\infty} P(s_{t'} | s_t, a_t) V(s_{t'}) \right]
	\label{eqn:bellman}
\end{equation}
where:
\begin{itemize}
	\item \textbf{Value Function $V(s_t)$}: Serves as a predictive model that evaluates the expected future performance of a portfolio from a given state $s_t$, representing current market conditions, asset valuations, and portfolio composition. By calculating the cumulative expected returns, the value function incorporates both immediate gains and future prospects, considering the probabilities of various market scenarios and their timings. It guides the RL agent in assessing the long-term effectiveness of different investment strategies, taking into account factors such as market volatility, portfolio diversification, and the risk-return balance. Essentially, $V(s_t)$ aids in algorithmic decision-making by projecting the future value of a portfolio under different conditions and strategies, facilitating the selection of optimal investment actions.
	\item \textbf{Immediate Reward $R(s_t,a_t)$}: Quantifies the instantaneous gain or loss resulting from an action $a_t$ taken in state $s_t$. It reflects the immediate financial impact of trading decisions in portfolio management, such as the realized profit and loss after executing a trade. Within reinforcement learning, the concept of reward shaping is crucial, involving the design of the reward function to align with specific investment objectives and risk profiles. For instance, a hedge fund targeting absolute returns might shape $R(s_t, a_t)$ based on log returns to emphasize direct profitability. Conversely, a strategy focused on risk-adjusted returns might utilize a reward function derived from the Sharpe ratio or its variants, thereby incorporating risk considerations. Effective reward shaping plays a pivotal role in guiding the RL agent toward investment strategies that align with investor goals, striking a balance between immediate returns and risk management.
	\item \textbf{Discount Factor $\gamma$}: Determines how the RL agent weighs immediate versus future rewards. In finance, it functions similarly to the discounting of future cash flows to their present value. A higher value of $\gamma$ indicates a stronger emphasis on long-term gains, mirroring a strategic long-term investment approach in finance. By adjusting $\gamma$, the RL model can be fine-tuned to prioritize either short-term gains or long-term investment objectives, aligning with the specific goals and risk tolerance of the investor.
	\item \textbf{Transition Probability $P(s_{t+1}|s_t,a_t)$}: Represents the probability of moving from the current state $s_t$ to a new state $s_{t+1}$ after taking action $a_t$. In the financial context, this is similar to the probability of market shifts resulting from specific investment actions. It captures the essence of market volatility and the inherent unpredictability in financial decision-making. This concept in RL reflects the need to account for the variable nature of markets, where each action can lead to multiple potential future scenarios, each with its own likelihood and financial implications.
	\item \textbf{Value of Subsequent State $V(s_{t+1})$}: Reflects the expected utility for future states resulting from current actions. In financial terms, it is comparable to projecting the performance of a portfolio into the future, taking into account the potential outcomes of present investment decisions. This aspect of RL encapsulates the idea of forward-looking analysis in portfolio management, where the focus is not only on immediate results but also on how current choices shape future financial landscapes. It highlights the importance of strategic planning and anticipatory decision-making in both RL and financial investment, considering the long-term implications of actions taken today.
\end{itemize}
Transitioning from the traditional state-value focus of the Bellman equation, shown in Equation \ref{eqn:bellman}, our approach in financial portfolio optimization embraces a broader perspective. In classical Reinforcement Learning, the concept of an optimal policy, denoted as \( \pi^{*} \), is central. This policy is typically formulated to maximize the expected cumulative reward over time, as shown in the standard RL formulation:
\begin{equation}
	\pi^{*} = \argmax_{\pi} \mathbb{E}_{\pi} \left[\sum_{t=0}^{\infty} \gamma^t R_t \right]
\end{equation}
Within the financial context, our emphasis shifts to identifying an optimal policy \( \pi^{*} \) that maximizes expected utility over an infinite horizon. This shift is encapsulated in the following formulation for Financial RL, where the optimal policy is interpreted as the \textit{optimal algorithmic investment strategy}:
\begin{equation}
	\pi^{*} = \argmax_{\pi}  \mathbb{E}_{\pi} \left[\sum_{t=0}^{\infty} \gamma^t U(r_t, \sigma_t, X_t, w_t)\right]
\end{equation}
In this formulation, \( \pi^{*} \) is not merely focused on maximizing immediate returns but integrates a broader spectrum of financial metrics, including returns (\(r_t\)), risk (\(\sigma_t\)), external market factors (\(X_t\)), and portfolio weights (\(w_t\)). This multi-faceted approach reflects the intricate nature of financial decision-making, where balancing risk, return, and adapting to market dynamics are paramount:
\begin{itemize}
	\item \textbf{Optimal Policy $\pi^{*}$}: Denotes the strategy that maximizes expected utility over an infinite horizon, which is crucial in financial contexts. In Financial RL, this is comparable to identifying the most effective algorithmic investment strategy, tailored to the dynamics and complexities of financial markets.
	\item \textbf{Expected Utility Maximization $\mathbb{E}_{\pi}$}: Represents the expectation of utility under policy $\pi$, quantifying the long-term effectiveness of an investment strategy. This concept aligns with our earlier discussion on maximizing expected outcomes in MPT, subject to real-world constraints as detailed in Equation \ref{eqn:multi_period_optimization_probabilistic}.
	\item \textbf{Recursive Utility Function $U(r_t, \sigma_t, X_t, w_t)$}: Integrates critical financial metrics at each time step $t$, including returns ($r_t$), risk ($\sigma_t$), external market factors ($X_t$), and portfolio weights ($w_t$). This function is pivotal in reflecting the nuanced trade-offs between risk and return, aligning closely with investor objectives and their specific financial goals.
\end{itemize}
In our previous discussions, we differentiated between standard Reinforcement Learning (RL) and Deep Reinforcement Learning (DRL), noting the integration of neural networks in the latter. Now, we delve deeper into this distinction, focusing specifically on the role of neural networks and Deep Learning architectures within DRL. These sophisticated computational frameworks are pivotal in advancing portfolio management strategies, offering enhanced predictive and adaptive capabilities far beyond the scope of traditional RL. The following discussion will delve into the complexities of these neural networks and Deep Learning architectures, emphasizing their critical role in financial portfolio optimization. Renowned for their efficacy, Deep Learning architectures form the backbone of DRL, processing complex, high-dimensional, and non-stationary financial data with unparalleled efficiency. The advent of diverse architectures such as Convolutional Neural Networks (CNNs) for image and time-series data, Recurrent Neural Networks (RNNs) for sequential data, and Transformers\footnote{Transformers have gained popularity not only in time series forecasting but also in language processing tasks. They are widely used in Large Language Models (LLMs), such as those popularized in ChatGPT by OpenAI, for tasks like text generation, translation, and summarization.} for complex text and language processing tasks, has revolutionized how large and noisy financial datasets are handled. These architectures are adept at extracting pertinent patterns and insights from a wide array of data types, ranging from numeric and textual to audio and visual formats, significantly broadening their applicability across diverse financial contexts. This adaptability is evidenced by their demonstrated superiority over traditional econometric models when applied to asset classes and sectors such as energy, real estate, equities, and exotic derivatives \cite{wang2019review, del2020energy, boukas2021deep, zheng2023deep, cao2021deep}. Furthermore, recent scientific advancements have extended Deep Learning's capabilities in long-term forecasting, challenging prevailing industry misconceptions \cite{caterini2018generic, cuomo2022scientific}. This progress in Deep Learning, underpinned by a generalized mathematical formulation of neural networks, lays the foundation for understanding their operational framework in financial portfolio optimization:
\begin{definition}[Generic Representation of Deep Neural Networks \cite{caterini2018generic, cuomo2022scientific}]
	A deep neural network with \(L\) layers can be mathematically represented as a composition of \(L\) functions \(f_i: E_i \times H_i \to E_{i+1}\), where \(E_i, H_i,\) and \(E_{i+1}\) are inner product spaces for each \(i \in [L]\). The state variables are denoted as \(x_i \in E_i\) and the parameters as \(\theta_i \in H_i\). The network's output for an input \(x \in E_1\) is defined by a function \(F: E_1 \times (H_1 \times \cdots \times H_L) \to E_{L+1}\), expressed as:
	\begin{equation}
		F(x; \theta) = (f_L \circ \cdots \circ f_1)(x),
		\label{eqn:generic_neural_net}
	\end{equation}
	where \(\theta\) encapsulates the parameter set \(\{\theta_1, \ldots, \theta_L\}\). This high-level formulation adapts to various architectures. Each layer function \(f_i\) boosts adaptability and learning, crucial for tasks like optimal portfolio allocations in RL. The head map \(\alpha_i\) and tail map \(\omega_i\) aid in understanding network behavior and facilitate backpropagation for neural training. In financial portfolio optimization, the final layer \(f_L\) might act as the strategic mastermind, comparable to an advanced trader, in an MDP or POMDP setting, positioning the neural network as the strategic core of these decision processes.
\end{definition}
In summary, our exploration of Reinforcement Learning (RL) in financial portfolio management has been comprehensive. Beginning with foundational concepts such as states, actions, and rewards, we progressed to understanding the recursive nature of decision-making through the Bellman equation. This journey led us through the intricacies of Markov Decision Processes (MDPs) and Partially Observable Markov Decision Processes (POMDPs), demonstrating their application in modeling dynamic financial markets. The integration of neural networks as computational engines within these frameworks was pivotal, enabling nuanced predictions essential for portfolio optimization. However, transitioning theoretical models to real-world financial markets presents challenges beyond technical complexities. It necessitates a blend of financial expertise, AI knowledge, and regulatory understanding to ensure system robustness, rigorous model validation, and compliance with regulatory and game theoretic constraints. While RL offers solutions beyond Modern Portfolio Theory (MPT), its application in finance is not without complexities and risks, demanding careful design and implementation to balance innovation with risk management. Having established a solid understanding of RL's theoretical and practical aspects in finance, we now turn to historical parallels between technology and finance. This exploration sets the stage for our novel sim-to-real frameworks, grounded in RL advancements, aiming to revolutionize portfolio management. Through empirical testing via simulations, we aim to validate our theoretical propositions and demonstrate the real-world viability of our innovative approaches.

\section{Advancing Portfolio Management: Novel Frameworks in Reinforcement Learning}

As we embark on presenting our novel frameworks in the realm of financial portfolio management, this section first establishes a foundational context, underscoring how modern finance has historically drawn inspiration from various scientific and technological disciplines. By delving into the intersections between technology and finance, we aim to provide a well-grounded rationale for our innovative approaches, rooted in the rich tapestry of interdisciplinary influences that have shaped financial strategies over time. Our exploration begins by highlighting the parallels between financial strategies and technological advancements, with a particular focus on recommendation systems. Both fields have navigated similar optimization challenges, employing Reinforcement Learning (RL) as a transformative tool. This convergence underscores RL's potential to significantly impact finance, a domain that has traditionally embraced methods from physics and probability theory, notably in areas such as options pricing and risk management \cite{akyildirim2014brief, weatherall2013physics}. Building on this historical context, we then introduce our novel \textit{simulation-to-real} (\textit{sim-to-real}) frameworks. These frameworks, inspired by advancements in robotics and mathematical physics, are specifically designed to address the complexities of modern financial portfolio management. They represent a strategic and innovative response, reflecting finance’s history of adapting cross-disciplinary methodologies. This is crucial given the high-stakes nature of and unique challenges faced by the financial sector. The final computational experiments aim to rigorously evaluate our \textit{sim-to-real} frameworks, gauging their potential impacts on the industry. Positioned at the cutting edge of current Artificial Intelligence research, these frameworks aim to stimulate discussion and the development of scalable, efficient, and practical solutions in finance. Our approach, rooted in historical precedents and cross-disciplinary insights, anticipates a future where finance is increasingly influenced by advanced  algorithms. In advocating for their responsible and effective application, we emphasize the need for these innovative technologies to be meticulously designed to meet the complex and highly regulated demands of modern international finance.

\subsection{Parallel Evolution: Reinforcement Learning in Finance and Technology}
Building on our earlier discussions of Modern Portfolio Theory (MPT) in finance, this subsection examines the striking similarities between optimization processes in finance and technology, especially when confronted with predefined constraints. In finance, Modern Portfolio Theory (MPT) aims to maximize returns by strategically allocating investments across a mix of assets such as stocks, bonds, and derivatives, balancing risk and potential gains. Each choice is subject to a myriad of real-world constraints including risk profiles, market conditions, and regulatory frameworks. In finance, the optimization challenge resembles the task encountered in technology's recommendation systems. These systems are designed to construct portfolios of digital content, encompassing a diverse range of offerings from products to movies and social media advertisements. Their objective is twofold. Firstly, to maximize expected revenue for the firm by strategically aligning offerings with user interests. Secondly, to enhance the user's experience by precisely tailoring recommendations. This user-centric approach aims to ensure that users are presented with content that resonates with their preferences, geographical locations, and personal interests. The success of these recommendation engines is measured not only by traditional engagement metrics like click-through rates and viewing time but also by the degree to which they fulfill user expectations and drive revenue growth. This meticulous strategy in content curation mirrors the principles of portfolio optimization in finance, where the goal is to balance risk and return to meet specific investment objectives.  The historical development of both finance and technology reveals profound methodological and mathematical parallels. In finance, numerical optimization methods under MPT are employed to derive the best risk-adjusted returns for portfolios. This mirrors technology's early recommendation systems, which used mathematical techniques such as matrix factorization to predict user preferences for digital content \cite{tuzhilin2008large, bell2007lessons, steck2021deep, xue2017deep}. Central to both domains is the concept of utility maximization. In finance, utility functions are employed to balance risk against return, aligning with investor objectives. Conversely, in technology, utility functions gauge user satisfaction, focusing on digital engagement and content preferences. Furthermore, both fields confront analogous challenges. The  \textit{cold start} problem in finance, characterized by initiating portfolio allocations without prior data (in the case of new financial products), is comparable to the challenge in technology of recommending content to new users who lack engagement history. Additionally, the issue of \textit{sparse rewards} is a common obstacle in both domains \cite{takara4411793deep, gao2023survey}. In finance, especially in contexts such as options trading, the concept of sparse rewards is manifested in the form of delayed financial outcomes. Until the expiry of an option or its exercise, the profit and loss (PnL) remains uncertain. This delay in realizing PnL can range from minutes to days or even months, depending on the nature of the investment. Similarly, in the realm of technology, particularly in recommendation systems, sparse rewards are seen as delayed monetary returns. The financial benefit for a technology firm materializes only when a user takes a definitive action, such as clicking on an advertisement or making a purchase from recommended products. Here too, the delay before seeing PnL can vary significantly, from immediate responses to prolonged periods of user engagement. These parallels in problem-solving structures and strategies between financial portfolio management and technology-based recommendation systems underscore a fundamental mathematical symmetry. This symmetry provides a logical foundation for hypothesizing the potential for Reinforcement Learning (RL) to transform finance, drawing on its successful applications in technology. This analysis not only reinforces the connection between these domains but also highlights the potential for cross-domain insights to inform future advancements in financial portfolio optimization.

For financial professionals, this parallel offers valuable insights. The evolution and resolution of these challenges in the technology sector, particularly through the adoption of advanced Reinforcement Learning techniques, suggest a potential roadmap for similar advancements in financial portfolio management. By acknowledging and learning from the technological sector’s approach to these analogous problems, finance professionals can envisage and prepare for potential revolutionary shifts in portfolio management strategies. For those within the finance sector who may harbor skepticism about the transformative potential of advanced algorithms, the evolution and successes of RL in technology present a compelling case.  As advancements in technology continued to evolve, the classical methods in both portfolio optimization and recommendation systems encountered significant limitations in addressing the dynamic and complex nature of their respective domains. This is particularly evident in portfolio optimization, where conventional methodologies rooted in Modern Portfolio Theory (MPT) often face challenges adapting to the realistic constraints and high-dimensional, noisy data characteristics of modern markets. The technology sector, in its quest for more adaptive and scalable solutions, increasingly embraced Deep Learning and Reinforcement Learning. RL's capability to learn from user interactions and dynamically adapt strategies proved to be evolutionary in addressing the intricate challenges of recommendation systems. The shift from static, algebraic solutions to dynamic, data-driven RL approaches indicates a potential evolutionary path for financial portfolio management. Notably, in the technology sector, issues once deemed computationally intractable have been effectively resolved by elite teams of mathematicians and physicists at major technology firms. These groups leverage innovative RL systems, achieving high efficiency, low latency, and scalability, thus enabling real-time recommendations for millions of users \cite{zhu2023pearl, bennett2021off, mehta2017review, zhang2019deep, afsar2022reinforcement, lin2023survey}.  

This historical parallel, where two distinct fields faced similar mathematical challenges and evolved towards RL-based solutions, strongly suggests that the future of finance is poised to be reshaped by these advanced algorithmic approaches \cite{mehta2017review, zhang2019deep, afsar2022reinforcement, lin2023survey, almahdi2017adaptive, sato2019model, aboussalah2020continuous, niu2022metatrader, zhang2020deep, felizardo2022reinforcement, ritter2017machine}. Much like how advanced algorithms revolutionized the advertising industry - a field once dominated by marketing and business experts - finance is poised to undergo a similar transformation. Sophisticated algorithmic solutions may challenge and reshape traditional practices, potentially altering the roles and tasks traditionally associated with investors. In envisioning a future increasingly driven by advanced algorithms in finance, we assert the need for prudence and sustainability in their application, especially given the stringent regulatory environment and the immense responsibility of managing client funds. Given the unique complexities of the financial sector and the responsibility of managing substantial funds, the deployment of Reinforcement Learning techniques must be conducted with meticulous care. Unlike recommendations in social media or e-commerce, errors in financial portfolio optimization can lead to risk exposures amounting to hundreds of millions or billions of dollars, with far-reaching consequences for investment firms, financial institutions, and sovereign wealth funds. Thus, the integration of RL into finance necessitates not only advanced algorithmic development but also a profound comprehension of market nuances and robust risk management protocols. Particularly in leveraged scenarios, the potential financial ramifications underscore the importance of rigorous risk assessment and stress testing practices. 

Collaboration is crucial in these practices, necessitating alignment among scientific, quantitative, risk, and regulatory teams to ensure RL's responsible utilization in high-stakes environments. Furthermore, academic researchers in this field are obliged to communicate assumptions and potential risks transparently. In integrating advanced technologies like Reinforcement Learning into financial portfolio optimization, effective collaboration among engineers, scientists, finance and legal professionals is paramount. This multidisciplinary synergy is vital, not only to drive innovation but also to ensure thoughtful application of RL within the highly regulated and complex landscape of finance. In the technology sector, the popular adage \textit{move fast and break things} encapsulates a culture that values rapid innovation and risk-taking. However, when considering the application of similar technological advancements in finance, this approach necessitates cautious reconsideration. A reckless importation of this ethos into the financial domain could lead to dire consequences, potentially sparking severe financial crises and triggering stringent regulatory responses, which could hinder future innovation. This paper advocates for a more deliberate and sustainable approach - \textit{move prudently and build sustainably} - as a guiding principle in the development of financial technology. Our aim is to spark and contribute to discussions about the future of financial technology, highlighting the need for a balanced approach that recognizes the unique risks and responsibilities of the financial sector. As we progress, we will delve into hypothetical case studies that offer simplified yet insightful perspectives on the unique challenges faced in finance as opposed to technology, emphasizing the importance of an interdisciplinary, cautious and collaborative approach.

\textbf{Practical Implications in Financial Portfolio Management:} In applying Reinforcement Learning to portfolio optimization, it's crucial for academic and industry research to transparently justify methodologies and reasoning. The diverse nature of financial markets and regulatory environments means that theoretically sound strategies might vary significantly based on specific market conditions and investor objectives. Consequently, the developers of such mathematical and computational artifacts must clearly articulate the assumptions underpinning the chosen approach and their respective rationale. This clarity is vital for understanding the real-world applicability and constraints of these sophisticated mathematical models in the complex landscape of financial portfolio management  and optimization\footnote{The development of RL systems for financial applications is an intricate endeavor that encompasses detailed mathematical modeling and considerable computational efforts. It demands not only expertise in algorithmic design but also adherence to stringent software engineering standards. Minor implementation errors can lead to significant operational complications, underscoring the need for precision in both the mathematical and software engineering facets of RL. This highlights the importance of creating robust, reliable, and efficient systems for high-stakes financial applications.}. To illustrate these points, we present a few simplified  open-ended questions, highlighting the significant complexity of building industry-grade RL systems in portfolio optimization and financial RL more broadly:
\begin{enumerate}
	\item  \textbf{Understanding Time Horizon \& Market Microstructure: } The transferability of an RL agent from one market domain to another poses significant challenges. Take, for instance, an RL agent (algorithm) developed for portfolio optimization in the energy derivatives market. The question arises: is it appropriate or even feasible to apply this agent to a disparate context such as a cross-asset strategy? Such a shift would involve transitioning from a market with specific characteristics to one with potentially vastly different microstructures and trading dynamics. This case underscores the need for domain-specific knowledge and strategy alignment. Moreover, the time horizon of the trading strategy plays a crucial role. An RL agent trained for high-frequency trading, where decisions are made on a millisecond basis, may not be suitable for long-term, long-only investment strategies. This distinction is rooted in the unique market microstructures and investment objectives associated with different trading styles and asset classes. It exemplifies the importance of understanding both the time horizon and the microstructural nuances of the market when deploying RL agents in finance \cite{biais2005market, easley2021microstructure, baldacci2021quantitative}.
	\item \textbf{Understanding Dynamical Systems \& Non-Stationarity:} The adaptability of RL agents to dynamic market conditions is put to the test in scenarios such as the onset of the COVID-19 pandemic. An agent trained in pre-pandemic market conditions might face significant challenges in adjusting to the subsequent, drastically altered market dynamics. This raises critical questions: has the agent been rigorously stress-tested to handle the volatility and shifts in market behavior seen post-COVID? Are the risk metrics utilized by the agent, sufficiently robust and adaptable to suit different investment styles and client risk profiles in this new market landscape? This scenario accentuates the inherent difficulty in dealing with non-stationary environments in financial markets, a problem that even state-of-the-art RL models grapple with. It underscores the importance of a comprehensive design of experiments \cite{mead2012statistical, parkinson2021we, schiefer2021statistical, kang2020design, munger2021statistical}, and continual adaptation and stress testing in financial RL applications. Such a meticulous approach is crucial for developing RL systems that are not only theoretically sound but also resilient to market evolution.
	\item \textbf{Understanding Asymmetric Risk \& Game Theory Dynamics:} A critical aspect of applying RL in finance is navigating asymmetric risk profiles and the game-theoretic nature of financial markets. This complexity is rooted in the constantly changing rules and dynamics that govern financial systems, unlike the more stable natural laws observed in the natural sciences. In the financial world, the introduction of new regulations, sudden market shocks, or strategic moves by major market players can drastically alter the market environment. This fluidity presents unique challenges for RL agents, which must be capable of understanding and responding to asymmetric risks and the strategic behavior of other market participants. It leads to a scenario where the data available for training and validation of these agents is not only limited but also rapidly becomes outdated as market dynamics evolve. This situation renders financial RL arguably more complex than many other domains, making the development of these respective systems among the most intricate. The crucial question then becomes: are the RL agents designed to account for and adapt to these asymmetric risks and game-theoretic complexities? \cite{galla2013complex, bratvold2011game, pandey2015investors}
\end{enumerate}
This discussion sheds light on the nuanced notion of \textit{fully autonomous} agents in finance, which, upon closer examination, appears somewhat of a misnomer. Finance, with its intricate web of risk management and regulatory compliance, necessitates a human-in-the-loop (HIL) approach, essential for overseeing and validating algorithmic decisions \cite{rajendran2021human, taylor2023reinforcement, correia2019human,  wu2021human}. Taylor et al. (2023) articulate this necessity well: ``\textit{[One should] not think of RL as a fully autonomous paradigm, but instead as an iterative learning and development process involving both learning algorithms and humans. While better algorithms may chip away at this assumption, full autonomy in terms of problem identification, construction, and deployment is unlikely in the near-future. Instead, we argue that it is critical to consider how humans and RL agents can work together to solve sequential decision tasks}'' \cite{taylor2023reinforcement}.  In concluding this section, we have emphasized the vital role of human insight and expertise in guiding and understanding the application of Reinforcement Learning (RL) in portfolio optimization and broader financial contexts.  Effective RL systems in finance must navigate a landscape filled with uncertainties that often extend beyond the scope of sophisticated statistical algorithms. The presented scenarios highlight this blend as critical for the successful and responsible application of RL in finance. 


\subsection{Simulation-to-Real Transfer in financial RL: Bridging the Reality Gap}
Drawing upon the successful application of Reinforcement Learning in high-risk domains like mathematical physics and robotics, we introduce our novel \textit{Simulation-to-Real} (\textit{Sim-to-Real}) frameworks for financial portfolio optimization. This advancement is underpinned by the interdisciplinary convergence of applied mathematics, computational science, and finance. These fields collectively reveal synergies and analogous methodologies, shaping our algorithmic solutions for financial Reinforcement Learning. Central to our approach is Sim-to-Real transfer learning, a method pivotal in the development of high-stakes, autonomous systems such as self-driving cars. Here, RL systems are exhaustively tested in simulated settings that closely mimic diverse real-world scenarios. This methodical simulation is critical in preparing the systems to navigate and adapt to the multifaceted and unpredictable nature of real-world environments, effectively narrowing the \textit{reality gap} – the often observed divergence between simulated models and real-world outcomes. In the realm of finance, the reality gap is characterized by the unique challenges posed by fluctuating market conditions, regulatory changes, and trading constraints. Our RL models for financial portfolio optimization are meticulously crafted to be resilient, adaptable, and progressively evolving in response to the dynamic nature of financial markets. We emphasize a strategic application of RL, deeply rooted in the practical realities of market dynamics and regulatory frameworks.  By integrating Sim-to-Real principles from sectors like robotics and mathematical physics, our goal is to significantly enhance the practicality and flexibility of RL in financial portfolio optimization. This approach not only aligns with our broader research objectives of fusing theoretical insights with real-world market complexities but also advocates for meticulous engineering, comprehensive domain expertise, and the responsible implementation of financial RL systems. Such a strategy echoes the critical insights gained from robotics, underscoring the importance and intricacies of deploying Sim-to-Real transfer in the financial sector:
\begin{quote}
\textit{"There was wide agreement that the ultimate goal is to design robotic systems that live in the real-world \dots A key aspect in sim-to-real transfer is the choice of simulation. Independently of the techniques utilized for efficiently transferring knowledge to real robots, the more realistic a simulation is the better results that can be expected \dots Reinforcement learning algorithms often rely on simulated data to meet their need for vast amounts of labeled experiences. The mismatch between the simulation environments and real-world scenarios, however, requires further attention to be put to methods for sim-to-real transfer of the knowledge acquired in simulation"} - Perspectives on sim2real transfer for robotics, H{\"o}fer et al (2020) \cite{hofer2020perspectives}.
\end{quote}
The advancements in high-risk robotics, as discussed by H{\"o}fer et al (2020), have been achieved through addressing significant technical challenges and innovations. As we apply sim-to-real constraints and methodologies similar to those in robotics to our financial RL models, we anticipate encountering comparable challenges. This underscores the need for innovation and meticulous application in finance, a domain with considerably high stakes. Our literature review on Reinforcement Learning and Artificial Intelligence in robotics has led to the formulation of three fundamental principles. These principles are not only grounded in advanced computational science but are also acutely aware of the complexities and challenges unique to financial markets. Our objective is to establish a preliminary framework tailored to the specific needs of financial portfolio management, while also adhering to the stringent regulatory frameworks that govern this sector. Drawing on insights from a wide array of fields, including robotics and mathematical physics \cite{abeyruwan2023sim2real, chen2021understanding, zhao2020sim, rusu2017sim, hu2022provable, bellemare2023distributional, abdar2021review, lockwood2022review, zhu2023uncertainty, bianchin2019secure, zhou2021multi, tobin2017domain, kontes2020high, vuong2019pick, lee2019network, foster2021offline, xie2021policy, xie2022role, wells2021explainable, madumal2020explainable, gajcin2024redefining, kormushev2013reinforcement, ibarz2021train, dulac2021challenges, zhu2020ingredients, singh2019end, cabi2019scaling, gupta2022unpacking, hadfield2017inverse, julian2020never, brunke2022safe}, we now introduce our novel \textit{Simulation-to-Real} (\textit{Sim-to-Real}) frameworks for financial RL. These frameworks are distinctively designed to address the specific challenges and nuances of financial RL use-cases:
\begin{enumerate}
\item \textbf{Realistic Reward Shaping \& Environmental Modeling:} Central to our sim-to-real transfer methods in high-risk financial applications is the precise construction of financial environments mirroring real-world market conditions. This requires in-depth modeling of transaction costs, market impact, and regulatory constraints, essential for effective RL strategies. Recent academic efforts, like using Generative Adversarial Networks (GANs) for market data generation, highlight the challenge of accurately replicating market dynamics. Without realistic simulation, such techniques risk being ineffective. The \textit{domain randomization} approach in robotics, where RL models face diverse simulated conditions to build resilience, is enhanced by incorporating concepts like \textit{concentrability}, which ensures training on a representative sample of the state space, and \textit{coverability}, which guarantees exposure to a wide range of market scenarios including rare but critical (black-swan) events. These additions are crucial for RL systems to effectively navigate and adapt to the volatile nature of financial markets. Reward shaping is equally crucial, requiring the selection of metrics such as log returns or the Sortino ratio, tailored to specific financial goals and investor profiles. This domain-specific knowledge is imperative for aligning RL agent actions with investor objectives. Furthermore, robust software engineering is key, particularly with open-source financial modeling libraries. The final RL software must be theoretically sound, resilient, and scalable, capable of operating efficiently in diverse computational environments, including industrial-grade investment settings.
\item \textbf{Robust Risk Analysis \& Statistical Stress Testing:} In the diverse and complex landscape of finance, it is imperative to design Reinforcement Learning systems that are specifically tailored to the unique characteristics of different asset classes and market microstructures, rather than attempting to create universally applicable models. This nuanced approach necessitates a comprehensive risk management strategy that incorporates conventional risk metrics, fine-tuned for particular asset classes and market sectors. A critical aspect of this approach is conducting extensive statistical analyses of performance metrics across a wide range of RL training scenarios. An active area of RL research also includes advanced causality-driven model explainability techniques, such as counterfactual explanations, which can significantly enhance understanding and trust in these systems. Emphasizing the importance of rigorous software engineering, especially for those transitioning from the computer science domain to financial applications, our methodology includes the utilization of advanced tools for serialization and cloud-based storage of performance data. This practice not only facilitates easy retrieval of crucial data during regulatory audits or client interactions but also ensures a high degree of transparency and accountability. By carefully considering these factors, we aim to develop RL systems that are not only robust in theoretical simulations but also adaptable and reliable in the real-world financial ecosystem, respecting the intricacies of each asset class and market microstructure.
\item \textbf{Strategic Resilience \& Fault Tolerance-Aware Engineering:} Drawing parallels from high-risk robotics, such as autonomous vehicle systems, financial RL systems require robust fault tolerance mechanisms to anticipate and effectively manage market extremities and external shocks. This entails engineering contingency protocols for scenarios where algorithms might malfunction or underperform, particularly during volatile market conditions. Leveraging strategies utilized in autonomous robotics to handle unpredictable events, financial RL systems need proactive design considerations for navigating market anomalies and systemic fluctuations. The burgeoning threat of adversarial attacks, particularly relevant for RL models processing sentiment data from fluctuating sources like social media, must be addressed. These models require rigorous vulnerability assessments to prevent potential exploitations that could adversely affect financial outcomes. A judicious approach might involve selectively excluding data sources where the risk-to-reward ratio is unfavorable. This proactive stance on system security and fault tolerance mandates exhaustive risk assessments, encompassing both traditional financial risks and those unique to algorithmic investments, such as data manipulation and model over-fitting. Integrating these comprehensive risk management strategies from the outset ensures the development of financial RL systems that are not only resilient but which also adhere to the stringent safety and adaptability standards observed in high-risk robotics applications.
\end{enumerate}
In our exploration of sim-to-real transfer methodologies from high-risk domains such as robotics to financial Reinforcement Learning (RL), we have established preliminary yet foundational guidelines. These guidelines are devised to guide the development of sophisticated RL strategies in the nuanced domains of portfolio optimization and broader financial RL. While they represent a significant initial contribution, it is essential to recognize that they are just the inception of a broader journey in this evolving field. These principles, we hope, will ignite ongoing discussions and foster safe yet innovative advancements, especially resonating with stakeholders in financial institutions, governments, and policymaking bodies worldwide. As the financial sector increasingly integrates complex algorithms, comprehensively understanding and addressing the multifaceted nature of these systems is vital. Our framework, while thorough in its current form, is not exhaustive and represents an initial step in aligning theoretical RL models with the dynamic and unpredictable nature of financial markets. With a focus on environmental realism, adaptability to market changes, and resilient system design, our work is committed to developing RL methodologies that can effectively meet the diverse challenges of the financial sector. Designed for both theoretical robustness and practical applicability, these methodologies reflect our dedication to pragmatic and effective solutions in financial management. As we conclude this part of our research, we approach a crucial stage; empirical validation of our theories through computational experiments with RL agents and our proprietary systems and market environments.

\section{Computational Experiments}

\begin{figure}[htp!]
	\centering
	\begin{tikzpicture}[node distance=5mm, font=\footnotesize]
		\node[draw, minimum width=10cm, minimum height=1cm, align=center] (train) {Train};
		\node[draw, minimum width=2.7cm, minimum height=1cm, align=center, right=of train] (gap) {Gap};
		\node[draw, minimum width=2.7cm, minimum height=1cm, align=center, right=of gap] (test) {Test};
		
		\draw [decorate,decoration={brace,amplitude=5pt},xshift=0pt,yshift=-3mm]
		($(train.south east) + (0,-0.1)$) -- ($(train.south west) + (0,-0.1)$)
		node [black,midway,yshift=-4mm] {\footnotesize 2017 to 2021};
		
		\draw [decorate,decoration={brace,amplitude=5pt},xshift=0pt,yshift=-3mm]
		($(gap.south east) + (0,-0.1)$) -- ($(gap.south west) + (0,-0.1)$)
		node [black,midway,yshift=-4mm] {\footnotesize 2022};
		
		\draw [decorate,decoration={brace,amplitude=5pt},xshift=0pt,yshift=-3mm]
		($(test.south east) + (0,-0.1)$) -- ($(test.south west) + (0,-0.1)$)
		node [black,midway,yshift=-4mm] {\footnotesize 2023};
		
		\draw[->] (train) -- (gap);
		\draw[->] (gap) -- (test);
	\end{tikzpicture}
	\caption{Experimental design setup: Our study employs Reinforcement Learning (RL) agents trained on historical data within our proprietary sim-to-real environments. In contrast to traditional backtesting approaches, our RL agents dynamically adjust portfolio weights across assets based on learned non-linear policies, offering a more adaptive and nuanced investment strategy driven by data. Evaluation is conducted over a single test year to assess the efficacy of the trained agents.}
	\label{fig:experimental_design}
\end{figure}
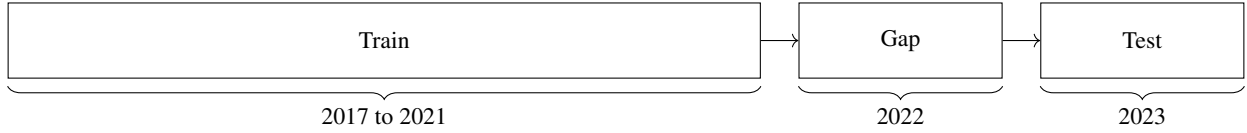

\textbf{Experiment Focus \& Evaluation Metrics:}  Our computational experiments aim to rigorously evaluate the novel \textit{simulation-to-real} framework, as delineated in previous sections. This framework is not only centered on applying state-of-the-art (SOTA) Reinforcement Learning algorithms, recognized for their effectiveness in literature under conditions of low or negligible transaction costs, but also on a comprehensive consideration of real-world financial factors. Our experimental design integrates key real-world factors, including market frictions and broker fees, surpassing traditional academic models to provide a realistic simulation of financial market conditions. It's important to note that certain implementation details are omitted due to the proprietary nature of our algorithms and system design. This led us to adopt a simplified experimental design that emphasizes a holistic, realism-driven approach to financial market simulations. We believe that this approach, while possibly affecting algorithmic performance, provides invaluable insights for both researchers and practitioners in industry and academia. As discussed in the previous sections, we posit that from a practitioner's standpoint, an algorithm rigorously stress-tested in realistic settings holds greater value, even if it demonstrates suboptimal performance. This is in contrast to algorithms that excel in academic research but may not withstand the complexities of real-world financial environments. Our work aims to bridge this gap, offering a framework that balances theoretical robustness with practical applicability.  Additionally, our experiment focus, is on specifically tailored to long-term horizon asset managers and market makers.  This distinction is vital, as the strategies and constraints relevant to these entities often differ markedly from those employed by independent retail proprietary traders or entities seeking pure alpha returns. By being transparent about this focus, we ensure that our experimental design aligns closely with the operational realities and regulatory considerations unique to these segments of the financial industry.  These experiments utilized high-performance cloud computing and advanced GPUs, conforming to industrial-grade software engineering practices. As part of our evaluation methodology, we incorporate a comprehensive set of performance metrics for asset-class  agnostic evaluation, as suggested by Lim, et. al (2019) \cite{lim2019enhancing} and Zhang et. al (2020) \cite{zhang2020deep}. These metrics collectively offer a comprehensive and nuanced view of portfolio performance, addressing various dimensions of risk and return, crucial for a holistic evaluation of any trading strategy:

\begin{table}[h]
	\centering
	\footnotesize
	\begin{tabular}{|p{0.1\linewidth}|p{0.35\linewidth}|p{0.46\linewidth}|}
		\hline
		\textbf{Metric} & \textbf{Definition} & \textbf{Explanation} \\ \hline
		$\E[R]$ & Annualized expected trade returns. & Indicates potential profitability and sets a baseline for performance expectation. \\ \hline
		$\text{std}(R)$ & Annualized standard deviation of trade returns. & Measures the volatility of returns, crucial for understanding the risk level of the strategy. \\ \hline
		DD & Annualized standard deviation of negative trade returns. & Captures the strategy's downside risk, critical for gauging potential losses. \\ \hline
		Sharpe & Sharpe ratio: $\E[R] / \text{std}(R)$. & Evaluates risk-adjusted returns, allowing comparison of performance against the risk taken. \\ \hline
		Sortino & Sortino ratio: $\E[R] / \text{DD}$. & Focuses on downside risk, offering a refined view of risk-adjusted performance. \\ \hline
		MDD & Maximum drawdown & Indicates the largest observed loss from peak to trough, essential for assessing strategy resilience. \\ \hline
		Calmar & Calmar ratio: $\E[R] / \text{MDD}$. & Relates annual return to maximum drawdown, providing a perspective on performance during adverse conditions. \\ \hline
		\% +ve  & Percentage of positive trade returns. & Reflects the strategy's consistency in yielding profits, a key aspect for investment decision-making. \\ \hline
		Avg+/Avg- & Ratio of average positive to negative trade \newline returns. & Shows the balance between average gains and losses, indicating overall trade effectiveness. \\ \hline
	\end{tabular}
	\caption{Performance Metrics for Algorithm Evaluation}
	\label{table:performance_metrics}
\end{table}
\textbf{Reward Function:} In evaluating performance, we adopted the Differential Sharpe Ratio (DSR) as our reward function. Articulated in the classic Financial RL paper by Moody and Saffell (1998), the DSR can be formulated as follows \cite{moody1998performance}:
\begin{equation}
	D_t = \frac{B_{t-1}\Delta A_t - \frac{1}{2} A_{t-1}\Delta B_t}{(B_{t-1} - A^2_{t-1})^{\frac{3}{2}}},
\end{equation}
where \(A_t\) and \(B_t\) represent the exponential moving averages of the returns and squared returns, respectively. \(\Delta A_t\) and \(\Delta B_t\) are the changes in these averages at time \(t\). This metric, utilizing exponential moving averages, is particularly suited for the dynamic and online nature of RL environments, offering immediate, risk-adjusted performance feedback vital for RL decision-making\footnote{The DSR's denominator, raised to the power of \( \frac{3}{2} \), emerges from its first-order expansion, capturing the immediate impact of current returns on the Sharpe Ratio. This formulation is unique to the DSR, specifically designed for online applications \cite{zhang2024discounted}.}. The DSR's focus on consistent returns and effective risk management aligns well with our goals of guiding the agent toward risk-adjusted strategies similar to those of a long-only asset manager. Different reward metrics might be preferred for strategies focusing on absolute returns or conservative approaches, emphasizing the need to tailor the reward function to specific investment objectives and regulatory requirements.

\textbf{Data Selection Rationale:} The data selection process for our study was meticulously planned to ensure both the validity and practical applicability of our models. We selected 11 high-volume U.S. equities and indexes from varied sectors including industrials, energy, financials, and consumer goods. This choice was driven by two main factors: Firstly, to minimize market impact and transaction costs. High-volume stocks were selected for their liquidity, which is crucial in real-world trading environments, reducing the potential market impact and costs associated with trading activities. Secondly, our focus was on enabling effective long-term horizon portfolio allocation strategies. These stocks and indexes are well-suited for strategies involving real-world Exchange Traded Funds (ETFs) and individual stocks. By opting for a low cardinality set of assets, as supported by existing literature, we aimed to focus on the sim-to-real dynamics within a controlled yet realistic market setting, avoiding excessive market frictions. Additionally, we started with a moderate initial cash amount and explicitly excluded the use of \textit{leverage} or borrowing, to further reduce market impact and align with real-world trading constraints. 

\textbf{Reinforcement Learning Models and Experimental Results:} We have carefully chosen a set of Reinforcement Learning (RL) algorithms that have been demonstrated to surpass traditional investment strategies, such as those based on Modern Portfolio Theory (MPT), as discussed in earlier sections\footnote{While the detailed technical intricacies of these algorithms are beyond the scope of this paper, the provided intuitive explanations aim to highlight their unique attributes and operational analogies in the context of financial portfolio management. It is vital to note the challenges inherent in replicating financial RL experiments, especially regarding computational resources. Studies from Wang et. al (2021) and others \cite{wang2021deeptrader, jiang2017deep, yu2019model, sood2023deep} highlight the importance of advanced computing power, vectorized training, and sophisticated memory buffer systems for scaling experiments and improving computational efficiency. Variations in hardware, software, and statistical experimentation design can lead to different outcomes, even with seemingly similar methodologies, features and model parameters.}. For clarity, especially for readers well-versed in finance terminology, we reiterate that the term \textit{policy} within this context refers to an \textit{algorithmic investment strategy}:
\begin{itemize}
	\item \textbf{Advantage Actor-Critic (A2C)}: Technically, A2C operates with a dual architecture, combining a policy model (the actor) and a value model (the critic). It simultaneously learns a policy and a value function, balancing immediate rewards with long-term value. This makes A2C efficient in high-dimensional action spaces. Intuitively, think of A2C as having a trader (actor) making decisions, with a risk management team (critic) evaluating and guiding these decisions for optimal outcomes.
	\item \textbf{AlphaOptimizerNet Simplified (AONS)}: Our proprietary AONS algorithm, which is based on more recent developments in the state-of-the-art (SOTA) Artificial Intelligence literature was originally designed for proprietary high-frequency intra-day research. The neural architecture was adapted and simplified for our long-term asset management experiments.
	\item \textbf{Proximal Policy Optimization (PPO)}: PPO is innovative in its approach to updating policies, using a clipped objective function to prevent large, destabilizing updates. It maintains a balance between exploring new strategies and exploiting known ones, crucial in unpredictable financial markets. Intuitively, PPO can be seen as an adaptive trader, constantly refining strategies based on market feedback and risk-return profiles.
	\item \textbf{Soft Actor-Critic (SAC)}: SAC employs a stochastic policy framework and off-policy learning, effectively exploring diverse strategies. This makes it adept at handling market uncertainty and volatility. Intuitively, SAC resembles a trader using probabilistic models to anticipate and adapt to various market conditions, enhancing strategy adaptability.
\end{itemize}

Table \ref{table:rl_performance_metrics} details the performance metrics for each RL agent during this out-of-sample backtesting phase, providing insights into their ability to navigate the complexities of the financial market landscape. Our out-of-sample backtesting yields a nuanced perspective on the applicability of reinforcement learning (RL) models under real-world market constraints. Although RL models are often touted for their potential to surpass classical investment strategies, our findings suggest that their efficacy might be compromised under realistic market conditions. These results preliminarily support our hypothesis that our novel sim-to-real considerations may be crucial for the practical deployment of RL models in financial environments, beyond being merely additive.
\begin{table}[htp!]
	\centering
	\caption{Out-of-Sample Backtesting Performance Metrics for RL Agents in Portfolio Optimization}
	\label{table:rl_performance_metrics}
	\begin{tabular}{|l|l|l|l|l|l|l|l|l|}
		\hline
		\textbf{Model} & \textbf{$\mathbf{E[R]}$} & \textbf{std(R)} & \textbf{DD} & \textbf{Sharpe} & \textbf{Sortino} & \textbf{MDD} & \textbf{Calmar} & \textbf{Avg+/Avg-} \\
		\hline
		A2C & $0.1382$ & $0.0152$ & $0.0151$ & $0.7927$ & $1.1738$ & $-0.2276$ & $0.5248$ & $-0.9826$ \\
		\hline
		AONS & $\mathbf{0.2309}$ & $0.0161$ & $\mathbf{0.0157}$ & $\mathbf{1.1459}$ & $\mathbf{1.7244}$ & $\mathbf{-0.2395}$ & $\mathbf{0.8291}$ & $\mathbf{-1.0373}$ \\
		\hline
		PPO & $0.1416$ & $0.0152$ & $0.0152$ & $0.8094$ & $1.1897$ & $-0.2511$ & $0.4876$ & $-0.9841$ \\
		\hline
		SAC & $0.0115$ & $0.0156$ & $0.0162$ & $0.1810$ & $0.2557$ & $-0.2776$ & $0.0360$ & $-0.9120$ \\
		\hline
	\end{tabular}
\end{table}
\begin{figure}[htp!]
	\centering
	\includegraphics[width=1\textwidth]{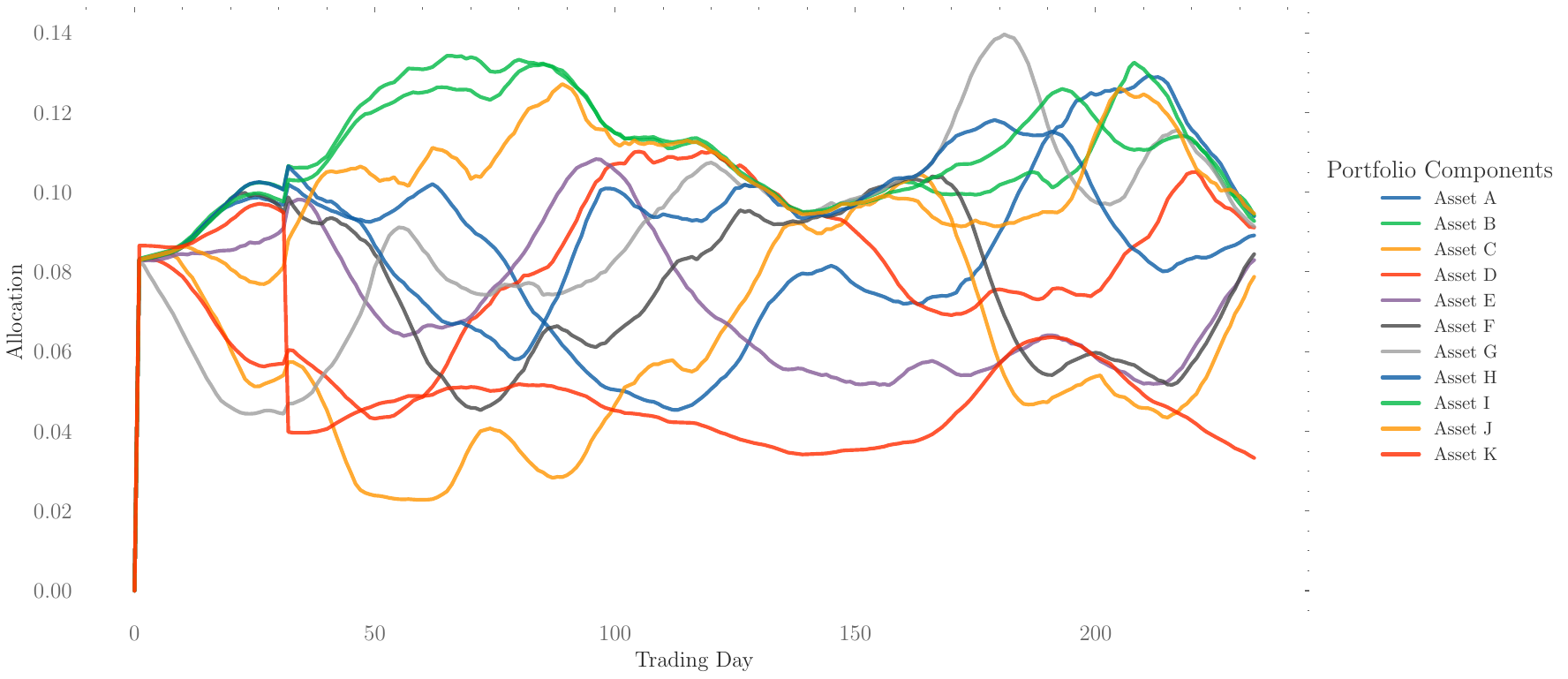}
	\caption{Dynamic portfolio allocation over time by our proprietary \textit{AlphaOptimizerNet}, suggesting a short-term momentum trading strategy. The model's allocation across multiple assets and its responsive adjustments to market conditions hint at its ability to discern meaningful financial patterns and implement complex strategies. Our empirical analysis validates this behavior. For deeper validation, we recommend an extensive examination using modern RL explainability algorithms, as detailed in our novel sim-to-real frameworks section.}
	\label{fig:aon_allocations_over_time}
\end{figure}
In particular, our proprietary \textit{AlphaOptimizerNet Simplified (AONS)} has demonstrated significant adaptability within recent market regimes. AONS posted an impressive 23.09\% return ($\E[R]$), indicating its strong performance in generating profits. This high return rate is noteworthy, especially when compared with other models like A2C, PPO, and SAC, as it highlights AONS's ability to adapt to market shifts and achieve superior returns. When considering risk-adjusted returns, AONS stands out with a Sharpe ratio of 1.1459 and a Sortino ratio of 1.7244. These metrics suggest that AONS efficiently balances returns with the risks it takes. Moreover, its Calmar ratio of 0.8291, despite a Maximum Drawdown (MDD) of -23.95\%, signals a strong recovery potential from losses. It's important to note that MDD, a crucial risk metric, is not annualized. Instead, it measures the largest single drop from peak to trough in the portfolio's value, offering insight into the downside risk over the entire investment period.  AONS's negative Avg+/Avg- ratio indicates an aggressive trading strategy, contrasting with the more passive approaches often found in classical Modern Portfolio Theory (MPT). This aggressive approach, focusing on capitalizing on market trends, is a significant departure from traditional strategies and highlights the dynamic nature of RL strategies in finance. Overall, while models like A2C, PPO, and SAC also achieved positive returns, they did not match AONS's high returns and risk-adjusted performance metrics. This underscores the effectiveness of RL in adapting to market changes and outperforming traditional single-period MVO solutions. Figure \ref{fig:aon_allocations_over_time} illustrates AONS's dynamic portfolio allocation strategy, exemplifying the model's proficiency in adjusting to short-term market trends. Such visualizations, coupled with an emphasis on model explainability, are crucial in a field where many financial AI studies tend to neglect the temporal evolution of trading strategies. Our replication attempts of these studies have revealed a trend where models often default to simpler strategies like Uniform Buy and Hold, emphasizing the importance of incorporating realistic market dynamics into financial AI research. Despite the promising results, they warrant cautious scrutiny, particularly in high-stakes financial domains. The aggressive stance of AONS may elicit concerns within traditional institutional risk management frameworks, reinforcing the need for judicious deployment of such models. The success of AONS, along with the positive outcomes from other RL models, calls for ongoing refinement and empirical validation to ensure their practical applicability in the ever-changing financial market landscape. Our study contributes to the discourse on integrating sophisticated AI techniques into portfolio management, advocating for a balanced approach that fuses technical ingenuity with pragmatic execution.
\begin{figure}[htp!]
	\centering
	\includegraphics[width=0.67\textwidth]{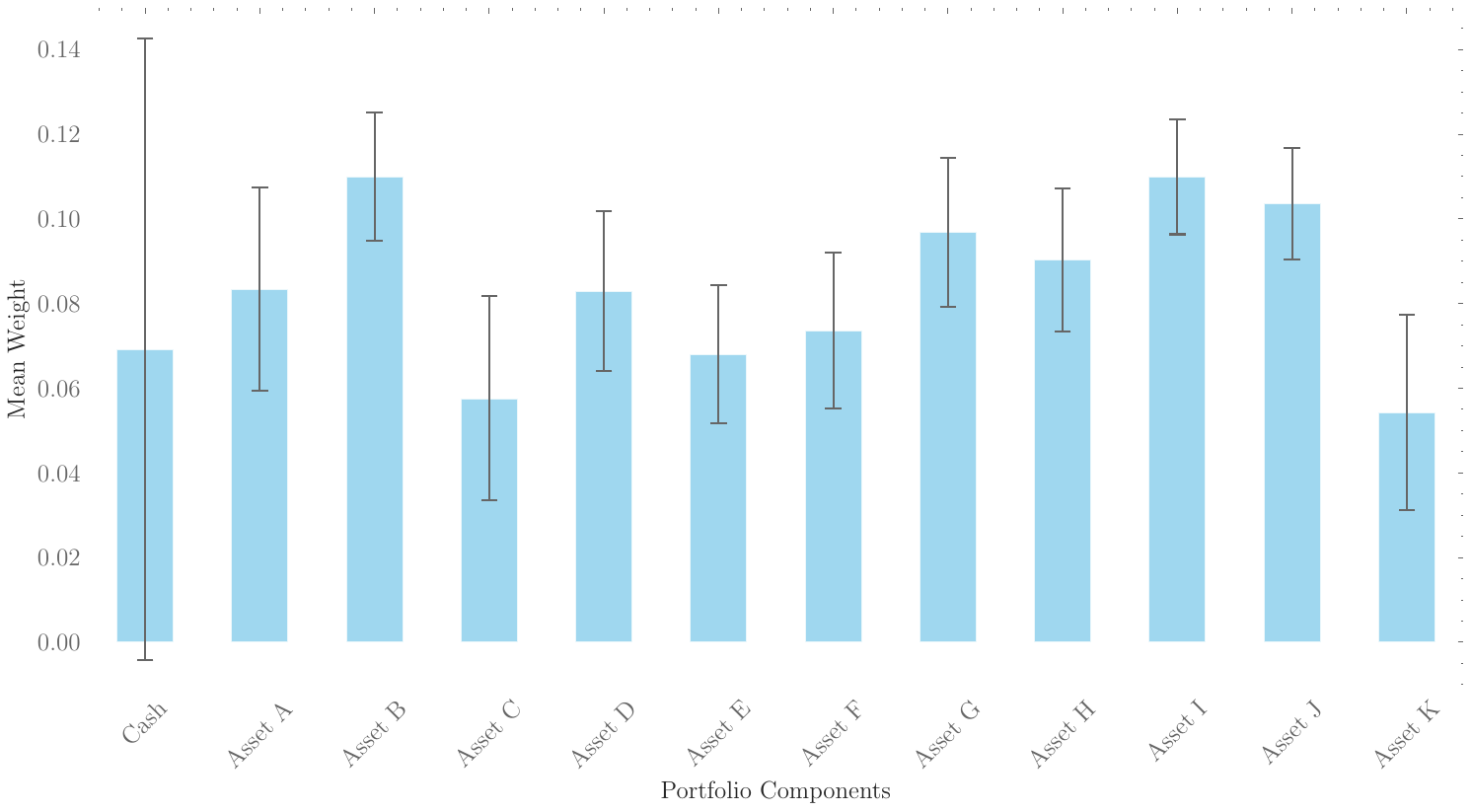}
	\caption{Average allocation per asset by our proprietary \textit{AlphaOptimizerNet} across the testing period, with error bars representing one standard deviation from the mean. This visual representation highlights the model's strategic distribution of investments, with the variance indicating its responsiveness to market conditions. The error bars may extend beyond the mean allocation to suggest the range within which the allocation varied over the period.}
	\label{fig:aon_mean_allocations}
\end{figure}

\section{Conclusion}

In conclusion, this paper marks a substantial step towards achieving our initial goals. We have successfully developed the \textit{AlphaOptimizerNet} Reinforcement Learning agent. Furthermore, we have explored the critical aspects of our novel sim-to-real-world transfer frameworks, contributing to the discourse on advanced AI applications in portfolio management. Our preliminary frameworks and extensive computational experiments are designed to serve as a cornerstone for future research and practical implementations in this field. We aspire for our findings and methodologies to elevate financial management standards and inform future explorations. Particularly for financial institutions managing substantial assets, our work offers a methodically designed reference point, potentially enhancing their wealth safeguarding strategies and ability to limit their exposure to risk. Our commitment to advancing this domain remains unwavering, as we continue to innovate in neural network architectures and refine our proposed frameworks. Our ultimate goal is to foster a financial ecosystem that is not only efficient and resilient but also attuned its broader impact on individual prosperity and collective economic growth. We hope that our current and future contributions will significantly \textit{advance investment frontiers}, bridging theoretical research and practical applications.

\textbf{Disclaimer:} The authors make no representation or warranty, express or implied, and disclaim all liability regarding the completeness, accuracy, or reliability of the information contained in this paper. This document is not intended as investment research or advice, nor as a recommendation, offer, or solicitation for the purchase or sale of any security, financial instrument, product, or service. 

\section*{References}
\setlength{\baselineskip}{0pt} 

{\renewcommand\MakeUppercase[1]{#1}%
	
	\printbibliography[heading=none, title=False]

\end{document}